\documentclass[manuscript,screen]{acmart}

\AtBeginDocument{%
  \providecommand\BibTeX{{%
    \normalfont B\kern-0.5em{\scshape i\kern-0.25em b}\kern-0.8em\TeX}}}

\setcopyright{acmlicensed}
\copyrightyear{2024}
\acmYear{2024}
\acmDOI{XXXXXXX.XXXXXXX}

\acmConference[XX]{XX}{XX}

\usepackage[marginal]{footmisc}
\usepackage{tabularx}
\usepackage{multirow}
\usepackage{alphabeta}
\usepackage{amsthm}
\usepackage{amsmath}
\usepackage{booktabs} 
\usepackage{threeparttable}
\usepackage{bbding}
\usepackage{xpatch}
\usepackage{pifont}
\usepackage{wasysym}

\usepackage{amssymb}
\usepackage{svg}
\usepackage{makecell}
\usepackage{amsmath,amsfonts}
\usepackage{algorithmic}
\usepackage{algorithm}
\usepackage{array}
\newcolumntype{C}[1]{>{\centering}p{#1}}
\usepackage[caption=false,font=normalsize,labelfont=sf,textfont=sf]{subfig}
\usepackage{textcomp}
\usepackage{stfloats}
\usepackage{url}
\usepackage{verbatim}
\usepackage{graphicx}
\usepackage{soul}

\newcommand{\cmmnt}[1]{}
\begin{document}

\title{Machine Unlearning: Taxonomy, Metrics, Applications, Challenges, and Prospects}


\author{Na Li} 
\affiliation{%
  \institution{Nanjing University of Science and Technology}
  \city{Nanjing}
  \country{China}
}
\affiliation{%
  \institution{Xidian University}
  \city{Xi'an}
  \country{China}
}

\author{Chunyi Zhou}
\affiliation{%
  \institution{Nanjing University of Science and Technology}
  \city{Nanjing}
  \country{China}
}

\author{Yansong Gao}
\affiliation{%
  \institution{CSIRO}
  \city{Canberra}
  \country{Australia}
  \postcode{ACT 2601}
  }
\email{garrison.gao@data61.csiro.au}

\author{Hui Chen}
\affiliation{%
  \institution{Nanjing University of Science and Technology}
  \city{Nanjing}
  \country{China}
}

\author{Anmin Fu}
\affiliation{%
  \institution{Nanjing University of Science and Technology}
  \city{Nanjing}
  \country{China}
}
\affiliation{%
  \institution{Xidian University}
  \city{Xi'an}
  \country{China}
}

\author{Zhi Zhang}
\affiliation{%
  \institution{University of Western Australia}
  \city{Perth}
  \country{Australia}
  \postcode{WA 6009}
  }
\email{zzhangphd@gmail.com}

\author{Shui Yu}
\affiliation{%
 \institution{University of Technology Sydney}
 \city{Sydney}
 \country{Australia}}
\email{shui.yu@uts.edu.au}


\begin{abstract}
Personal digital data is a critical asset, and governments worldwide have enforced laws and regulations to protect data privacy. Data users have been endowed with the `right to be forgotten' of their data. In the course of machine learning (ML), the forgotten right requires a model provider to delete user data and its subsequent impact on ML models upon user requests. 
Machine unlearning emerges to address this, which has garnered ever-increasing attention from both industry and academia. 
While the area has developed rapidly, there is a lack of comprehensive surveys to capture the latest advancements. Recognizing this shortage, we conduct an extensive exploration to map the landscape of machine unlearning including the (fine-grained) taxonomy of unlearning algorithms under centralized and distributed settings, debate on approximate unlearning, verification and evaluation metrics, challenges and solutions for unlearning under different applications, as well as attacks targeting machine unlearning. The survey concludes by outlining potential directions for future research, hoping to serve as a guide for interested scholars.
\end{abstract}




\maketitle

\section{Introduction}
Driven by an explosion of data and computational power, deep learning (DL) has showcased stunning performance in various applications such as self-driving \cite{Ndikumana21,Yurtsever20}, predicting a protein's 3D structure from its amino acid sequence \cite{Jumper21}, deciphering the genetic code and unveiling the secrets of hidden DNA diseases \cite{Cheng23}, and the very recent artificial intelligence generated content (AIGC) wave represented by text generation via ChatGPT \cite{Else23,Walker23}, image and video generation via diffusion model \cite{Yang22}, and code generation via Codex \cite{Chen21Evaluating}. These models are trained on user-contributed data \cite{Ma23}. 
Unintentionally, this raises privacy concerns as the model memorizes users' private information permanently, which might be leaked through known e.g., membership inference, property inference, and preference profiling attacks as well as yet disclosed privacy attacks.

By recognizing the importance of protecting user data privacy, national governments have issued a number of regulations including
EU's General Data Protection Regulation (GDPR) \cite{GDPR18}, Canada's Consumer Privacy Protection Act (CPPA), and America's California Consumer Privacy Act (CCPA) \cite{CCPA18}. These regulations provisions mandatory means of collecting, storing, analyzing, and utilizing personal data from citizens by related data consumers or organizations. Enforced by the `right to be forgotten', data consumers have to comply promptly with user requests to erase their data and eliminate any related impact. This endows data contributors' control over their data even after data release, fostering a willingness to share and contribute high-quality data. This, in turn, benefits (model) service providers by enabling higher service profits and reducing legal risks \cite{Cao15}.

Notably, forgetting data not only safeguards privacy in compliance with legal requirements but is also beneficial in other scenarios. It can unlearn adverse effects due to harmful data (e.g., adversarial data, poisoned data \cite{Zhang22Poison}, noisy labels \cite{Northcutt21}) or outdated data, thereby enhancing model security, responsiveness, and reliability. Furthermore, by unlearning victim data targeted by the adversary, it can mitigate a multitude of privacy attacks such as membership inference attacks and model inversion attacks, preventing sensitive training data private information leakage from the model.

\begin{figure*}
\centering
\includegraphics[scale=0.55]{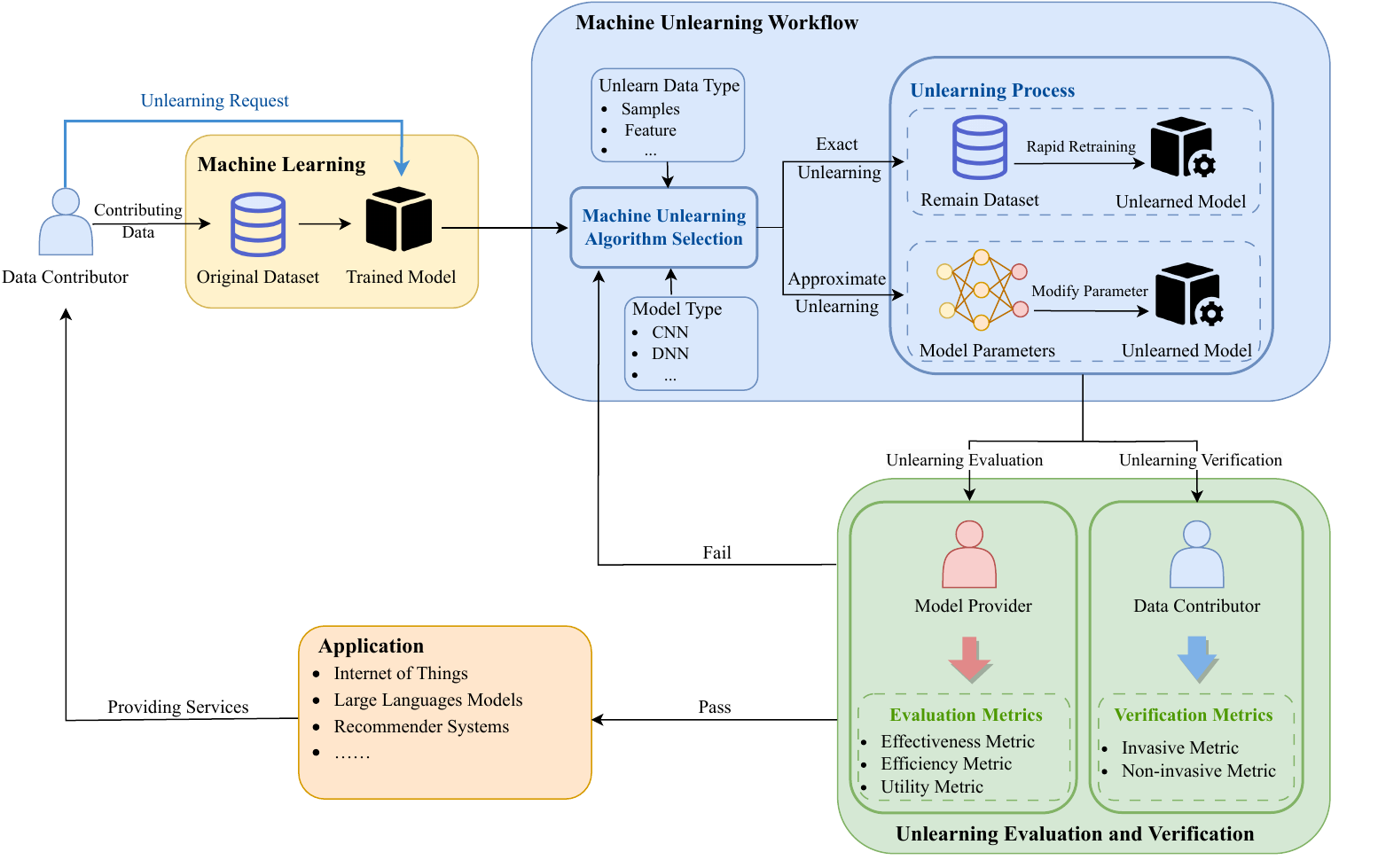}
\caption{Framework of Machine Unlearning.}
\label{fig:framework}
\end{figure*}

In the DL context, merely removing the raw training data from back-end databases is relatively meaningless. Because the DL model is still (explicitly or in-explicitly) memorizing the ingrained patterns and features that often contain sensitive details of requested data samples \cite{Arpit17,Carlini19}. Therefore, raw data related latent representation appears to be a challenging for being directly eliminated from the DL models. Existing conventional privacy protection techniques fail to meet these requirements, leading to an emergence of research direction in machine learning known as \textit{Machine Unlearning} (MU). The MU enables data contributors to actively withdraw their data used for model training, aiming at erasing its influence from the trained models as if it never existed, without compromising the model utility (as shown in Fig. \ref{fig:framework}).

Machine unlearning is undoubtedly becoming more relevant given the endowed `right to be forgotten'. The most straightforward approach is to periodically retrain a new model on the remaining dataset (without the cohort of data to be unlearned) from scratch. 
Unfortunately, this obviously renders costly computational overhead as well as response latency, especially for ever-increasing dataset size and model complexity. 
This training-from-scratch is more likely to be unacceptable for model providers and users and, thus not practical for real-world applications. 
Machine unlearning attempts to overcome the severe shortcomings of the aforementioned naive approach. Existing MU methods can be divided into two main categories based on whether there is a necessity of (re)training operations on the remaining dataset, which are exact unlearning and approximate unlearning. Exact unlearning aims at expediting the (re)training process, while approximate unlearning obviates the need for retraining by directly altering the model parameters, both making the model after unlearning indistinguishable from one obtained using the naive approach.

\subsection{Contributions of This Survey}

Machine unlearning is undergoing rapid development, but there exists a notable lack of comprehensive summaries and analyses to better depict the state-of-the-art. For instance, there is a deficiency in addressing the challenges of machine unlearning across various applications, as well as a lack of security analyses on machine unlearning. These shortcomings have prompted us to undertake a thorough investigation. 
This survey covers key research in machine unlearning from 2015 to 2024, including the taxonomy of unlearning algorithms under centralized and distributed settings---latter is often overlooked, evaluation and verification metrics, unlearning enabled applications, as well as attacks targeting threatening MU. 
The aim of the survey is to provide a knowledge base that will promote further scholarship and innovation in this burgeoning MU field. The key contributions of this survey are summarized as follows:

\begin{itemize}
\item{We conducted a comprehensive review of existing machine unlearning algorithms for diverse tasks including large language models, systematically categorizing them by unlearning mechanisms, and critically analyzed the merits and limitations inherent to each (sub)category.}

\item{We provided a detailed analysis of the challenges faced by machine unlearning in distributed learning settings, systematically categorizing its methods and comparing their advantages and disadvantages.}

\item We devised a taxonomy for the existing verification and evaluation metrics in machine unlearning. This taxonomy aims to assist both data owners and model owners, emphasizing the primary focus of each metric.

\item We underscored the diverse applications of machine unlearning in various scenarios, emphasizing its distinct advantages in optimizing models and defending against security and privacy attacks. This flexibility allows for rapid deployment and adaptation in real-world situations tailored to specific requirements.

\item We conducted a thorough examination of the challenges associated with machine unlearning, outlining potential research directions for future scholars to explore and reference.

\end{itemize}

\subsection{Comparison with Existing Surveys}

\begin{table*}[]
\caption{Comparison with Current Surveys }
\label{table:surveys}
\scalebox{0.65}{
\renewcommand{\arraystretch}{1.5}
\begin{threeparttable}
\begin{tabular}{|l|l|cc|ccc|ccc|cc|cllll|ccc|}
\hline
\multirow{1}{*}{\textbf{Paper}} & \multirow{1}{*}{\textbf{Year}} & \multicolumn{2}{l|}{\textbf{Verification Metrics}}  & \multicolumn{3}{c|}{\textbf{Evaluation Metrics}}   & \multicolumn{3}{c|}{\textbf{Taxonomy}}         & \multicolumn{2}{c|}{\textbf{Distributed Unlearning}}                  & \multicolumn{5}{c|}{\textbf{Applications}}     & \multicolumn{3}{l|}{\textbf{Attacks against MU}}  \\ \cline{3-20}

&     &\makebox[0.02\textwidth][l]{ Inva.\tnote{1}} &Non-inva.\tnote{2} & Effec.\tnote{3} & Effic.\tnote{4} & \makebox[0.04\textwidth][l]{Utili.\tnote{5}} &Exac.\tnote{6} &Appr.\tnote{7} &Deba.\tnote{8} &\makebox[0.1\textwidth][r]{Challenges}   
&FU \tnote{9}& LLMs\tnote{10} &RES\tnote{11} & IoT\tnote{12} &PD \tnote{13}& \makebox[0.04\textwidth][l]{AD\tnote{14}} &\makebox[0.04\textwidth][r]{MIA\tnote{15}} &\makebox[0.04\textwidth][r]{DPA\tnote{16}} &\makebox[0.07\textwidth][l]{OUA\tnote{17}}  \\\hline

 \cite{Tahiliani2021}       & 2021           &\makebox[0.05\textwidth][r]{\XSolid}     &\makebox[0.02\textwidth][l]{\XSolid}  & \makebox[0.03\textwidth][c]{\XSolid}     &\makebox[0.03\textwidth][c]{\XSolid}   &\makebox[0.03\textwidth][l]{\XSolid}       &\makebox[0.02\textwidth][c]{\XSolid}     &\makebox[0.02\textwidth][c]{\XSolid} 
 &\makebox[0.02\textwidth][c]{\XSolid}
 &\makebox[0.07\textwidth][c]{\XSolid}     &\makebox[0.02\textwidth][l]{\XSolid}   &\XSolid     & \XSolid    &\XSolid   & \XSolid  & \XSolid & \makebox[0.03\textwidth][r]{\XSolid}     &\makebox[0.03\textwidth][c]{\XSolid}   &\makebox[0.05\textwidth][l]{\XSolid}    \\ \hline
 
 \cite{Mercuri2022}  & 2022     &\makebox[0.05\textwidth][r]{\XSolid}     &\makebox[0.02\textwidth][l]{\XSolid}   & \Checkmark     &\Checkmark    &\makebox[0.03\textwidth][l]{\XSolid}    & \Checkmark    &\Checkmark &\XSolid    &\XSolid &\XSolid       & \XSolid    & \XSolid  & \XSolid    & \XSolid    & \XSolid     & \makebox[0.03\textwidth][r]{\XSolid}     &\XSolid    &\makebox[0.05\textwidth][l]{\XSolid}\\ \hline
 
 \cite{Nguyen2022}    & 2022    &\makebox[0.05\textwidth][r]{\XSolid}     &\makebox[0.02\textwidth][l]{\XSolid}  & \makebox[0.03\textwidth][c]{\XSolid}     &\makebox[0.03\textwidth][c]{\XSolid}   &\makebox[0.03\textwidth][l]{\XSolid}      &\makebox[0.02\textwidth][c]{\Checkmark}    &\makebox[0.02\textwidth][c]{\Checkmark} &\makebox[0.02\textwidth][c]{\XSolid}   &\makebox[0.03\textwidth][c]{\XSolid}  &\makebox[0.02\textwidth][l]{\XSolid}    & \XSolid        & \Checkmark                  & \XSolid                    & \XSolid            & \XSolid & \makebox[0.03\textwidth][r]{\Checkmark}     &\makebox[0.03\textwidth][c]{\Checkmark}   &\makebox[0.05\textwidth][l]{\XSolid}  \\ \hline
 
\cite{Zhang23review}     & 2023    &\makebox[0.05\textwidth][r]{\XSolid}     &\makebox[0.02\textwidth][l]{\XSolid}   & \makebox[0.03\textwidth][c]{\Checkmark}     &\makebox[0.03\textwidth][c]{\Checkmark}   &\makebox[0.03\textwidth][l]{\XSolid}       & \makebox[0.02\textwidth][c]{\Checkmark}    &\makebox[0.02\textwidth][c]{\Checkmark}     &\makebox[0.02\textwidth][c]{\XSolid}  &\makebox[0.03\textwidth][c]{\XSolid}   &\makebox[0.02\textwidth][l]{\XSolid}      & \XSolid         &\XSolid                    &\XSolid                      & \XSolid             & \XSolid          & \makebox[0.03\textwidth][r]{\XSolid}     &\makebox[0.03\textwidth][c]{\Checkmark}   &\makebox[0.05\textwidth][l]{\XSolid}     \\ \hline
 
 \cite{Qu_2023}        & 2023    &\makebox[0.05\textwidth][r]{\XSolid}     &\makebox[0.02\textwidth][l]{\XSolid}   & \makebox[0.03\textwidth][c]{\XSolid}     &\makebox[0.03\textwidth][c]{\XSolid}   &\makebox[0.03\textwidth][l]{\XSolid}      & \makebox[0.02\textwidth][c]{\Checkmark}    &\makebox[0.02\textwidth][c]{\Checkmark}   &\makebox[0.02\textwidth][c]{\XSolid}  
 &\makebox[0.03\textwidth][c]{\XSolid}   &\makebox[0.02\textwidth][l]{\XSolid}    & \XSolid   &\XSolid     & \XSolid    &\XSolid   & \XSolid    & \makebox[0.03\textwidth][r]{\XSolid}     &\makebox[0.03\textwidth][c]{\XSolid}   &\makebox[0.05\textwidth][l]{\XSolid} \\ \hline
 
\cite{Xu2023}        & 2023    &\makebox[0.05\textwidth][r]{\XSolid}    &\makebox[0.02\textwidth][l]{\XSolid}  & \makebox[0.03\textwidth][c]{\XSolid}     &\makebox[0.03\textwidth][c]{\XSolid}   &\makebox[0.03\textwidth][l]{\XSolid}      & \makebox[0.02\textwidth][c]{\Checkmark}    &\makebox[0.02\textwidth][c]{\Checkmark}  &\makebox[0.02\textwidth][c]{\XSolid}  &\makebox[0.03\textwidth][c]{\XSolid}   &\makebox[0.02\textwidth][l]{\XSolid}     & \XSolid   & \Checkmark    &\XSolid   & \XSolid   & \XSolid & \makebox[0.03\textwidth][r]{\Checkmark}     &\makebox[0.03\textwidth][c]{\Checkmark}   &\makebox[0.05\textwidth][l]{\XSolid}  \\   \hline

  \cite{yang2023survey}   & 2023     &\makebox[0.05\textwidth][r]{\XSolid}     &\makebox[0.02\textwidth][l]{\XSolid}   & \makebox[0.03\textwidth][c]{\XSolid}     &\makebox[0.03\textwidth][c]{\Checkmark}   &\makebox[0.03\textwidth][l]{\XSolid}       &\makebox[0.02\textwidth][c]{\XSolid}     &\makebox[0.02\textwidth][c]{\XSolid}     &\makebox[0.02\textwidth][c]{\XSolid} 
  &\makebox[0.03\textwidth][c]{\Checkmark}
  &\makebox[0.02\textwidth][l]{\Checkmark}      & \XSolid        & \XSolid                  & \XSolid                     & \XSolid            &\XSolid      & \makebox[0.03\textwidth][r]{\XSolid}     &\makebox[0.03\textwidth][c]{\XSolid}   &\makebox[0.05\textwidth][l]{\XSolid}       \\ \hline  

 \cite{liu2023survey}   & 2023     &\makebox[0.05\textwidth][r]{\XSolid}     &\makebox[0.02\textwidth][l]{\XSolid}   & \makebox[0.03\textwidth][c]{\XSolid}     &\makebox[0.03\textwidth][c]{\XSolid}   &\makebox[0.03\textwidth][l]{\XSolid}       &\makebox[0.02\textwidth][c]{\XSolid}     &\makebox[0.02\textwidth][c]{\XSolid}    &\makebox[0.02\textwidth][c]{\XSolid}  
 &\makebox[0.03\textwidth][c]{\Checkmark} 
 &\makebox[0.02\textwidth][l]{\Checkmark}      & \XSolid        & \XSolid                  & \XSolid                     & \XSolid            &\XSolid      & \makebox[0.03\textwidth][r]{\XSolid}     &\makebox[0.03\textwidth][c]{\XSolid}   &\makebox[0.05\textwidth][l]{\XSolid}       \\ \hline 

  \cite{shaik2023exploring}    & 2023     &\makebox[0.05\textwidth][r]{\XSolid}     &\makebox[0.02\textwidth][l]{\XSolid}   & \makebox[0.03\textwidth][c]{\XSolid}     &\makebox[0.03\textwidth][c]{\XSolid}   &\makebox[0.03\textwidth][l]{\XSolid}       &\makebox[0.02\textwidth][c]{\XSolid}     &\makebox[0.02\textwidth][c]{\XSolid}   &\makebox[0.02\textwidth][c]{\XSolid} &\makebox[0.03\textwidth][c]{\rotatebox{90}{\XSolid}}    &\makebox[0.02\textwidth][l]{\XSolid}      & \XSolid        & \Checkmark                  & \XSolid                     & \XSolid            &\XSolid      & \makebox[0.03\textwidth][r]{\XSolid}     &\makebox[0.03\textwidth][c]{\Checkmark}   &\makebox[0.05\textwidth][l]{\XSolid}       \\ \hline

\cite{Xu24}        & 2024     &\makebox[0.05\textwidth][r]{\XSolid}     &\makebox[0.02\textwidth][l]{\XSolid}   & \makebox[0.03\textwidth][c]{\Checkmark}     &\makebox[0.03\textwidth][c]{\Checkmark}   &\makebox[0.03\textwidth][l]{\XSolid}      &\makebox[0.02\textwidth][c]{\XSolid}     &\makebox[0.02\textwidth][c]{\XSolid}   &\makebox[0.03\textwidth][c]{\XSolid}    &\makebox[0.02\textwidth][c]{\XSolid} &\makebox[0.02\textwidth][l]{\XSolid}     & \XSolid        & \XSolid        &\XSolid       & \XSolid     &    \XSolid   & \makebox[0.03\textwidth][r]{\XSolid}     &\makebox[0.03\textwidth][c]{\XSolid}  &\makebox[0.05\textwidth][l]{\XSolid}\\ \hline
 
 \textbf{Ours} & \textbf{2024}      & \makebox[0.05\textwidth][r]{\Checkmark}    &\Checkmark  &\Checkmark     &\Checkmark    &\makebox[0.03\textwidth][l]{\Checkmark}   &\Checkmark    &\Checkmark &\Checkmark &\Checkmark    &\Checkmark    & \Checkmark  & \Checkmark    & \Checkmark     & \Checkmark & \Checkmark  & \makebox[0.03\textwidth][r]{\Checkmark}     &\Checkmark    &\makebox[0.05\textwidth][l]{\Checkmark}\\ \hline
 \end{tabular}
 \begin{tablenotes}
        \footnotesize
        \item(1) Inva. -- Invasive Metric  
        \quad(2) Non-inva. -- Non-invasive Metric  
        \quad(3) Effec. -- Effectiveness Metric  
        \quad(4) Effic. -- Efficiency Metric  
        \quad(5) Utili. -- Utility Metric
        \quad(6) Exac. -- Exact Unlearning 
        \quad(7) Appr. -- Approximate Unlearning\\
        \quad(8) Deba. -- Debate on Approximate Unlearning
        \quad(9) FU -- Federated Unlearning  
        \quad(10) LLMs -- Unlearning for Large Language Models  
        \quad(11) RES -- Unlearning for Recommender Systems 
        \quad(12) IoT -- Unlearning for Internet of Things 
        \quad(13) PD -- Passive Defense
        \quad(14) AD -- Active Defense  
        \quad(15) MIA -- MU-specific Membership Inference Attack
        \quad(16) DPA -- MU-specific Data Poisoning Attack 
        \quad(17) OUA -- Over-unlearning Attack 
      \end{tablenotes}
  \end{threeparttable}
  }
 \end{table*}

Table \ref{table:surveys} summarizes the differences between our survey and previous surveys. The main differences lie in several aspects.
First, we clearly distinguish and comprehensively clarify verification and evaluation metrics, which are vital in paving the way for the practical implementation of machine unlearning. 
In contrast, many existing surveys confuse these metrics or even overlook them.
Second, we provide a detailed summary of the debate on approximate unlearning, which is missing from other surveys.
Third, we delve into the challenges of machine unlearning in the distributed setting, analyzing current methods in detail, which is a topic not sufficiently addressed in prior surveys. 
Fourth, we point out the difficulties faced by machine unlearning in different application scenarios (e.g., large language models) and how machine unlearning can be turned into a defense against various attacks. 
Previous surveys often lack depth and comprehensiveness in this analysis. Finally, we categorize various emerging attacks that threaten machine unlearning elaborating on their malicious purposes and assumptions. This facet is often addressed only peripherally in most surveys, merely offering analyses of a select subset of attack methodologies.

\subsection{Survey Organization}
This survey is organized as follows. Section \ref{2} presents preliminaries of machine unlearning. Section \ref{5} discusses verification and evaluation metrics for measuring the quality of machine unlearning. Section \ref{3} categorizes existing machine unlearning algorithms, delving into each fine-grained category and thoroughly analyzing its respective strengths and weaknesses. Section \ref{4} investigates the emerging machine unlearning in distributed settings. Section \ref{6} highlights potential applications e.g., erasing harmful information enabled by machine unlearning. Section \ref{7} summarizes existing privacy and security attacks targeting machine unlearning. Lastly, section \ref{8} delineates the current challenges confronting machine unlearning and posits auspicious directions for prospective research.

\section{Preliminaries}
\label{2}

Machine unlearning is dedicated to facilitating trained models to selectively erase data points and their impact on the model, thereby safeguarding personal data privacy. This section will furnish a comprehensive introduction to machine unlearning, encompassing its definition, the requisite properties that algorithms need to fulfill, and its operational workflow.

\subsection{Definition}
Let $\mathcal{D} = \{ ({x_{\rm{i}}},{y_i})\} _{i = 1}^n$ represents the original training dataset, where $x_{i}$ represents the training data point and  $y_{i}$ represents the label corresponding to  $x_{i}$. There exists a learning algorithm, denoted by a function \textsf{A}($\cdot$), so the model that completes training on  $\mathcal{D}$ can be denoted as \textsf{A}($\mathcal{D}$). Let $\mathcal{D}_u \subset \mathcal{D}$ represent the subset of $\mathcal{D}$ requested to be removed. The complement of this subset, $\mathcal{D}_r$ ($\mathcal{D}_r \cap \mathcal{D}_u = \emptyset$ and  $\mathcal{D}_r \cup \mathcal{D}_u = \mathcal{D}$), represents the remaining dataset. The function $\bar {\textsf{A}}( \cdot )$ encapsulates the machine unlearning algorithm.
If a request for unlearning is initiated, it is imperative to erase all information associated with $\mathcal{D}_u$ from the trained model $\textsf{A}(\mathcal{D})$. This is accomplished by executing $\bar{\textsf{A}}(\cdot)$, which constructs an unlearned model denoted as $\bar{\textsf{A}}(\mathcal{D}_u, \textsf{A}(\mathcal{D}))$. The unlearned model is expected to be indistinguishable from the retrained model, expressed as $\textsf{A}(\mathcal{D}_r)$ \cite{Cao15}, which is obtained through naive retraining from scratch.

\begin{table*}[]
\centering
\caption{Summary of Notations}
\label{table:notation}
\setlength{\tabcolsep}{4mm}{
\begin{tabular}{l|l|l|l}
\hline
\textbf{Notation}   & \textbf{Description}                            & \textbf{Notation} & \textbf{Description}          \\ \hline
$\mathcal{D}$ & The original training dataset            & \textsf{A}($\cdot$)  & Machine learning algorithm       \\
$\mathcal{D}_u$ & Dataset request to be removed      & $\bar {\textsf{A}}( \cdot )$   & Machine unlearning algorithm\\
$\mathcal{D}_r$ & The remaining training dataset         &\textsf{A}($\mathcal{D}$) & The original trained model  \\ 
$w$        & The original model parameter                &\textsf{A}($\mathcal{D}_r$)   &The retrained model\\
$w_r$     & The unlearned model parameter    &$\bar{\textsf{A}}(\mathcal{D}_u, \textsf{A}(\mathcal{D}))$ & The unlearned model\\
$x_{i}$        & The training data points    &$r_t$       & Number of mini-batches \\
$y_{i}$        & Labels corresponding to $x_{i}$   &$b$  & The optimal noise\\ 
\hline
\end{tabular}}  
\end{table*}

\subsection{Properties}
A well-designed machine unlearning algorithm should fulfill the following four properties: 
\begin{itemize}
\item{\textit{Effectiveness}: In machine unlearning, an unlearned model is obtained by selectively forgetting data points from an original model possessing robust generalization capabilities. The crucial requirement is that the unlearned model must completely remove any information associated with the forgotten data, rendering it as if the model has never been exposed to this data.}
\item\textit{Efficiency}: The unlearning needs to respond promptly to user requests and be completed within the legally required time frame. In this context, the unlearning algorithm should be computationally cost-efficient.
\item\textit{Utility}: Post unlearning, the utility e.g., the accuracy of the model on the remaining dataset should be consistent with its capability before unlearning. The unlearning process should not impair the utility of the model.
\item\textit{Compatibility}: The designed unlearning algorithm should be easily deployable on the existing machine learning models.
\end{itemize}

\subsection{Workflow}
The general workflow of machine unlearning is depicted in Fig.~\ref{fig:framework}. In the context of Machine Learning as a Service (MLaaS), a model has been trained on an original dataset to provide high-performance services. Due to privacy and security concerns, data contributors may want to withdraw his/her personal data  (or specific data features, etc.) from the deployed model \cite{Cao15}. 
Consequently, data contributors initiate an unlearning request to the model provider (the server), demanding that the model effectively forget the corresponding data as if the model had never used it \cite{Cao18Efficient}.
Upon receiving the unlearning request, the server selects an appropriate unlearning algorithm (e.g., sub-categorical methods in exact unlearning or approximate unlearning) to execute an unlearning based on specific factors such as unlearning data type (e.g., text, image, graph), model type (e.g., Convolutional Neural Networks, Deep Neural Networks, Graph Neural Networks, Linear models), and requirements prioritizing either thorough deletion or speed \cite{Schelter20Amnesia,Schelter21HedgeCut}. 
Then, the server obtains an unlearned model that does not contain any information about the forgotten data. 

\section{Verification and Evaluation Metric}
\label{5}
Upon unlearning, model providers assert that they have removed the influence of forgotten samples $\mathcal{D}_u$ from their models. However, malicious providers may breach this claim for reasons such as resource conservation, interest in sensitive user data, data theft, and avoiding performance degradation \cite{He21}. To address this, verification metrics are crucial for data contributors to assess whether the unlearned model still contains relevant $\mathcal{D}_u$ information. As outlined in Section \ref{2}, idea machine unlearning algorithms should satisfy the properties of effectiveness, efficiency, and utility. Model providers need evaluation metrics aligning with these properties to assess their unlearning algorithms, while data contributors require verification metrics to verify genuine data removal.

\begin{figure}[htbp]
\includegraphics[scale=0.8]{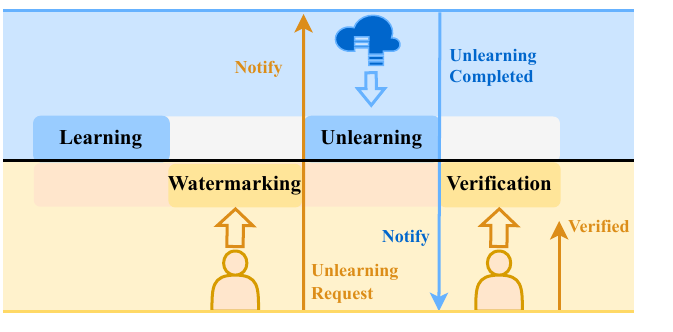}
\caption{Invasive Verification Process.}
\label{fig:active}
\end{figure} 

\subsection{Verification Metrics}
Verification metrics are used by the data contributor to verify whether the unlearned model has indeed completely unlearned $\mathcal{D}_u$ as claimed by model providers. According to whether the verification impacts the model training, it is classified into invasive and non-invasive metrics.

\subsubsection{Invasive Metric}
Invasive metrics have two key steps: watermarking and verification (as shown in Fig. \ref{fig:active}) \cite{gao2022verifi}. In the watermarking step, data contributors employ a watermarking method to watermark the data (e.g., $\mathcal{D}_u$, model parameters) that need to be checked. In the verification step, data contributors actively analyze the unlearning result on the watermarks immediately after model providers claim that unlearning is completed. It is crucial to note that the invasive verification process must be carried out stealthily, without model providers aware that data is being watermarked, to avoid any malicious attempts to reduce the accuracy of the verification result. 

\begin{itemize}
\item\textbf{Watermark-based Metric.}
Watermarking embeds backdoor triggers or owner-specific binary strings into the data (e.g., $\mathcal{D}_u$, model parameters) \cite{Sommer20,Izzo21}, as well as create adversarial examples to capture the unique properties of the model to assist unlearning verification  \cite{gao2022verifi}.

\begin{itemize}
\item\textbf{Backdoor-based Watermark.} Sommer \textit{et al.} \cite{Sommer20} conducted a verification through the backdoor. The process begins with the watermarking step, in which data contributors design backdoor patterns that change the predictions of $\mathcal{D}_u$ to an artificial target label. The poisoned samples (watermarks) are then uploaded to the server. During the verification step, once the model has completed the unlearning, data contributors evaluate the statistics on the backdoor success rate. If the model retains the relevant information of the $\mathcal{D}_u$, the backdoor success rate will exhibit notably high. Gao \textit{et al.} \cite{gao2022verifi} extended \cite{Sommer20} to unlearning verification under federated learning settings. However, \cite{Sommer20, gao2022verifi} just randomly selects pixels and sets their values of 1 as the backdoor trigger, which can be easily detected by model providers. To counteract this, Guo \textit{et al.} \cite{guo2024ver} used the Least Significant Bit (LSB) algorithm to inject triggers into watermarks.

\item\textbf{Feature-based Watermark.} Izzo \textit{et al.} \cite{Izzo21} proposed the Feature Injection Test (FIT) mainly for the unlearning verification of linear classifiers. The watermarking step is where they inject a strong signal into $\mathcal{D}_u$, involving the addition of an extra feature which is set to zero. The training process will assign a significantly different weight from zero to this feature. The verification step is that, because unlearning will cause the weight on this feature to become zero again, so unlearning can be verified by detecting the change of the unlearned model's weight for this special feature before and after deletion. The closer the change is to zero the more successful the unlearning is.

\item\textbf{Adversarial Example-based Watermark.}
Gao \textit{et al.} \cite{gao2022verifi} introduced the first verification metric under federated learning, watermarking the local model of a particular user to verify his entire data is removed from the global model. Since the unique nature of a Deep Neural Networks (DNN) classifier can be represented by its classification boundary \cite{cao2021ipguard}, they leveraged the boundary fingerprinting \cite{cao2021ipguard} to find decision boundary fingerprints as watermarks for uniquely identifying the local model. Specifically, starts with a watermarking step, data contributors produce adversarial examples as watermarks near the decision boundary to characterize the robustness of the local model. Then fine-tune the local model to obtain a watermarked local model that demonstrates a more smoothed (robust) boundary around the watermarks. The watermarked local model is then transmitted to the server for aggregation into the global model. The final step is the verification, which checks the prediction results of the global model on adversarial watermarks. Given that genuine unlearning is expected to rapidly diminish the smoothed (robust) boundary around the watermarks, the wrong model's predictions indicate successful unlearning.
\end{itemize}
\end{itemize}

\subsubsection{Non-invasive Metric}
A common scenario is that data contributors do not actively watermark data in advance to verify model providers's claimed authenticity---e.g., in legacy models. Meanwhile, invasive metrics may raise security concerns, and the accuracy of verification results can sometimes be degraded by defensive invasive actions. In this case, data contributors can adopt non-invasive metrics, which utilize the model's output or require model providers to offer cryptographic proof to verify the unlearning. Below we summarize existing non-invasive metrics.

\begin{itemize}
\item\textbf{Membership Inference Metric.} 
Membership inference can determine whether a given data sample exists in the training dataset \cite{Shokri17,Kurmanji23}. This helps verify whether the model still contains relevant information about $\mathcal{D}_u$. Graves \textit{et al.} \cite{Graves21} used membership inference in unlearning verification. They first trained a shadow model to mimic the unlearned model, then trained a binary meta-classifier with the shadow model's output (e.g., predictive probability or confidence) on both training and non-training data as meta-data.
Finally, this meta-classifier determines whether $\mathcal{D}_u$ is still in the training dataset using a similar format of meta-data from the unlearned model. Liu \textit{et al.} \cite{Ma23} calculated the Forgetting Rate (FR) by measuring the precision of the inference of the members before and after unlearning. The FR offers an intuitive measure of the success rate of member inference on $\mathcal{D}_u$ that changes from member to non-member after unlearning. When the FR value is 1, it indicates that $\mathcal{D}_u$ have been completely changed from member to non-member and unlearning is successful.
\item\textbf{Data Reconstruct Metric.}
Data reconstruction can reverse a model's training data information upon model outputs (in black-box) or parameters (in white-box) \cite{Graves21,Ye22}. 
In \cite{Graves21,Chundawat23Zero-Shot}, the unlearning is conducted at class-level rather than sample-level. To verify class-level data removal, they adopted a modified version of the threat model proposed by \cite{fredrikson2015model} to reconstruct the data's class information.
After initializing the input vector with zeros, a small amount of noise is injected. The threat model is optimized by gradient descent using the loss calculated concerning this input and the target class (forgotten class). 
After every $n$ steps of gradient descent, an image processing step is performed, which helps in the recognition of the generated images. The whole process is repeated for a certain number of epochs to get the final inverted images. If the unlearning has been done correctly, the data reconstructed from the unlearned model's parameters should not contain any information about the forgotten class. 
Salem \textit{et al.} \cite{Salem2019Updates-Leak} utilized the change in the output of a black-box model before and after unlearning to carry out an adversary attack that could reconstruct the unlearned sample to verify sample-level data removal.
\item\textbf{Cryptographic-based Metric.}
As most existing verification frameworks lack theoretical guarantees, Eisenhofer \textit{et al.} \cite{Eisenhofer2022Verifiable} constructed the first protocol for verifiable unlearning. This instantiation protocol uses succinct non-interactive argument of knowledge-based verifiable computation for proving model updates induced by unlearning, and hash chains for proving non-members in a training dataset. Model providers can use this protocol to provide cryptographic proof that an agreed-upon unlearning process has been executed. 
\end{itemize}

\begin{table*}[]
\centering
\caption{Summary of Metrics}
\label{table:metrics}
\scalebox{0.7}{
\renewcommand{\arraystretch}{1.5}
\begin{tabular}{|l|l|l|l|l|}
\hline 
\multicolumn{3}{|l|}{\textbf{Metrics}}  & \multicolumn{1}{l}{\textbf{Advantages}}   & \multicolumn{1}{|l|}{\textbf{Drawbacks}}  \\
\hline

\multirow{5}{*}[-3.0ex]{Verification Metrics}&\multirow{2}{*}[2.0ex]{Invasive Metrics}& \begin{tabular}[c]{@{}l@{}}Watermark-based \\ \cite{Sommer20, Izzo21, gao2022verifi, guo2024ver}\end{tabular}  &Conclusion can be obtained intuitively      &Damage model's performance  
\\
\cline{2-5}
&\multirow{3}{*}[-4.0ex]{Non-invasive Metrics}
&\begin{tabular}[c]{@{}l@{}}Data reconstruct  \\ \cite{Graves21,Ye22,Chundawat23Zero-Shot,fredrikson2015model,Salem2019Updates-Leak}\end{tabular}    &Conclusion can be obtained intuitively    &The implementation is complex 
\\
\cline{3-5}
&&\begin{tabular}[c]{@{}l@{}}Cryptographic-based  \\ \cite{Eisenhofer2022Verifiable}\end{tabular}   & Theoretical guarantees for verification     & Difficult for users to understand 
 \\\cline{3-5}
 
&&\begin{tabular}[c]{@{}l@{}}Membership inference \\ \cite{Shokri17,Kurmanji23,Graves21}\end{tabular}     & Inferences can take a variety of ways   &The implementation is complex  \\\hline

\multirow{6}{*}[-10.0ex]{Evaluation Metrics}
&\multirow{4}{*}[-9.0ex]{Unlearning Effectiveness}
&\begin{tabular}[c]{@{}l@{}}Relearn time \\ \cite{Chundawat23Zero-Shot,Golatkar21,Tarun21}\end{tabular} &  Evaluation processes easy to understand &Evaluation capability is weak \\
\cline{3-5}

&&\begin{tabular}[c]{@{}l@{}}Theory-based \\ \cite{Cao15,Bourtoule21,Guo20}\\\cite{Golatkar21,Golatkar20Eternal,Neel21,Sekhari21,Wu20}\end{tabular}  &Theoretical guarantees for evaluation  &Only for specific unlearning method
\\\cline{3-5}

&&\begin{tabular}[c]{@{}l@{}}Similarity-based \\ \cite{Golatkar20Forgetting,Golatkar21,Wu20,Graves21,Izzo21,Thudi22Unrolling,Chundawat23Zero-Shot}\end{tabular}  &Evaluation process is relatively rapid  &Need to compare with retrained model  \\\cline{3-5}

&&\begin{tabular}[c]{@{}l@{}}Accuracy on $\mathcal{D}_u$ \\ \cite{Brophy21,Wu20,Guo20,Izzo21,Schelter21HedgeCut}\\ \cite{Yan22ARCANE,Wu22,Graves21,Bourtoule21,Warnecke23}\end{tabular}  &Evaluation processes easy to implement  &The capability for evaluation is weak \\\cline{2-5}

&Unlearning Efficiency  &\begin{tabular}[c]{@{}l@{}}Unlearn speed \\ \cite{Brophy21,Cao15,Wu20,Schelter21HedgeCut,Yan22ARCANE}\\ \cite{Graves21,Izzo21,Warnecke23,Zhang22Prompt,Wu23GIF}\\ \cite{Bourtoule21,Chen22Graph,Guo20} \end{tabular}        &Evaluation processes easy to understand   &Need to compare with retrained model  \\
\cline{2-5}
&Model Utility & \begin{tabular}[c]{@{}l@{}}Accuracy on $\mathcal{D}_r$ \\  \cite{Cao15,Graves21} \end{tabular}        &Evaluation processes easy to implement   &Evaluation results are not convincing   \\\hline
\end{tabular}}
\end{table*}

\subsection{Evaluation Metrics}
Evaluation metrics are useful for model providers who can assess the effectiveness, utility, and efficiency of their unlearning.

\subsubsection{Effectiveness Metric}
Unlearning effectiveness means that the unlearned model should not contain any information about $\mathcal{D}_u$, as if the model had never seen the $\mathcal{D}_u$. By evaluating the effectiveness, model providers can measure the quality of the model's unlearning capabilities.

\begin{itemize}
\item {\textbf{Relearn Time Metric.}}
The relearn time is the number of epochs required for the unlearned model to regain the same accuracy on $\mathcal{D}_u$ as the original model when $\mathcal{D}_u$ is removed. It indirectly measures the amount of information residing in the unlearned model. If the relearn time is small, there is a high probability that the unlearned model retains more information. Otherwise, the unlearned model is close to the model that has never seen $\mathcal{D}_u$, so unlearning is effective \cite{Golatkar21,Tarun21}.  Chundawat \textit{et al.} \cite{Chundawat23Zero-Shot} indicated the unlearned model can rapidly regain significant accuracy but may not converge to the original accuracy on $\mathcal{D}_u$ for a long period. Hence, determining relearn time solely based on reaching or surpassing the original accuracy in epoch number can be misleading. To counter this, an α$\%$ margin around the original accuracy is introduced to calculate the Anamnesis Index (AIN) \cite{Chundawat23Zero-Shot}, defined as:

\begin{equation}
AIN = \frac{{{{\rm{r}}_t}(\bar{\textsf{A}}(\mathcal{D}_u, \textsf{A}(\mathcal{D})),{\textsf{A}(\mathcal{D})},\alpha )}}{{{{\rm{r}}_t}({\textsf{A}(\mathcal{D}_r)},{\textsf{A}(\mathcal{D})},\alpha )}}\
\end{equation}   

Notations can be recalled in Table \ref{table:notation}. AIN approaching 1 indicates effective unlearning, while values much lower than 1 suggest there is information left, corresponding to a shorter relearn time. AIN substantially exceeding 1 may, however, signify notable parameter changes, resulting in over-unlearning, potentially causing a `Streisand effect' (private data is more accessible to the adversary).

\item \textbf{Similarity-based Metric.}
The similarity between the unlearned model and the retrained model is illustrated by measuring the distance between them in terms of activation, weight, and distribution (the guarantee of the unlearning effectiveness provided becomes stronger in the above order). The higher the similarity, the better the unlearning.

\begin{itemize} 
\item\textbf{Activation Distance.} Activation similarity provides a weaker evaluation of unlearning. In \cite{Golatkar20Forgetting,Golatkar21}, the $L_{1}$-norm distance is utilized to quantify the similarity of the final activations between the unlearned and retrained model. The smaller $L_{1}$-norm is, the higher the unlearning effectiveness.
\item\textbf{Weight Distance.} As the modification of model weights is a common operation in most approximate unlearning processes, various similarity metrics are used to verify the correlation of weights between unlearned and retrained models. The common metrics are $L_{2}$-norm distance and cosine similarity. 
The smaller $L_{2}$-norm distance, or cosine similarity, the better unlearning effectiveness \cite{Wu20,Graves21,Izzo21}. Note that the cosine similarity is limited to the evaluation of classification tasks. Though these metrics are simple, they impose practical limitations such as the necessity to retrain the model from scratch and the potential deviation due to floating point operations. To counter this, an extended version of the $L_{2}$-norm distance metric for verifying approximate unlearning was proposed by Thudi \textit{et al.} \cite{Thudi22Unrolling}. They eliminate the necessity for naive retraining by calculating the $L_{2}$-norm distance between the final approximate unlearning weight and the initial weight, significantly reducing computational costs. 

\item\textbf{Distribution Distance.} Ensuring distribution similarity between the unlearned and retrained model provides a higher guarantee of unlearning efficacy while measuring such similarity is non-trivial. The KL divergence is a common metric used to measure the distribution distance between two models: the closer to 0, the better. Chundawat \textit{et al.} \cite{Chundawat23Bad} used Jensen-Shannon (JS) divergence to calculate Zero Retrain Forgetting (ZRF), which compares the output distribution before and after unlearning.
\end{itemize}

\item\textbf{Accuracy on $\mathcal{D}_u$ Metric.}
A well-trained model tends to have good generalization with high accuracy, especially for data in the training dataset. This property can be exploited to indirectly verify the effectiveness of unlearning \cite{Brophy21,Wu20}. For $\mathcal{D}_u$, the ideal accuracy should be the same as a model trained without seeing $D_{u}$ \cite{Schelter21HedgeCut,Yan22ARCANE}. Despite simple and straightforward, accuracy lacks sensitivity to the complexities and subtleties of the unlearning process. Nonetheless, the majority of studies have used this metric as the most basic form of evaluating unlearning effectiveness \cite{Brophy21,Wu20,Schelter21HedgeCut,Yan22ARCANE,Bourtoule21,Guo20,Graves21,Graves21,Izzo21,Warnecke23}.

\item {\textbf{Theory-based Metric.}}
Some machine unlearning algorithms e.g., certified unlearning are inherently designed to have properties that prove their effectiveness.
\begin{itemize}
\item\textbf{Retrain-based.} Exact unlearning is typically aimed at rapidly retraining with $\mathcal{D}_r$, which ensures that the distribution of an unlearned model and a naive retrained model are indistinguishable. For example, in \cite{Cao15,Bourtoule21}, When unlearning $\mathcal{D}_u$ is requested, only relevant data shards that contain the $\mathcal{D}_u$ are retrained, substantially minimizing computational costs for effectiveness. Note that this metric is inapplicable for approximate unlearning.
\item\textbf{Certified-based.} 
Certified unlearning was first proposed by Guo \textit{et al.} \cite{Guo20}, which adds delicate noise to the weights \cite{Golatkar21,Golatkar20Eternal,Neel21,Sekhari21,Wu20} or the loss function \cite{Guo20} based on differential privacy (DP). This guarantees that the outputs of the unlearned model are indistinguishable from the retrained model. However, this metric does not apply to models with non-convex loss functions that are common in deep learning.
\end{itemize}
\end{itemize}

\subsubsection{Efficiency Metric}
The efficiency of the unlearning process also needs to be verified, because `the right to be forgotten' requires the service provider to complete the unlearning operation within a specified time. The faster it is completed, the sooner the service can continue to be provided to users.
\begin{itemize}
\item {\textbf{Unlearn Speed Metric.}}
Unlearn speed (running time) can be used to assess the unlearning efficiency, measuring the time difference between unlearning and naive retraining. The larger the difference is, the quicker the system is at restoring privacy, security, and usability \cite{Cao15}. It can be observed from Table \ref{table:metrics} that the majority of studies utilize this metric to evaluate the unlearning efficiency.
\end{itemize}

\subsubsection{Utility Metric}
Utility means that the predictive accuracy of the model on the $\mathcal{D}_r$ should be consistent before and after unlearning. Removing data from the trained model may deteriorate its performance, which is undesirable for normal cases. Therefore, it is imperative to evaluate the performance of the unlearned model to ensure it is usable.

\begin{itemize}
\item {\textbf{Accuracy on $\mathcal{D}_r$ Metric.}}
For the $\mathcal{D}_r$ or test dataset, the ideal accuracy should be the same as a model trained without unlearned. This metric can be used to evaluate the influence of unlearning on remaining data and verify whether the utility of the unlearned model is adversely impacted. 
\end{itemize}

\noindent\textbf{Summary:} Verification metrics allow data contributors to verify that the unlearned model is genuinely accomplished and does not contain any traces of their data. Invasive verification metrics are relatively easy to implement with higher accuracy. 
However, they rely on watermarked data beforehand that can intrude into the model training process, which may impact the model's accuracy and essentially pose a security risk (e.g. backdoor-based watermarks could be maliciously exploited). 
Non-invasive verification metrics do not impact the model as they only passively use the model's output information, but the verification process is relatively complex, especially for membership inference where an additional binary meta-classifier needs to be trained (i.e., could be costly). 

Evaluation metrics facilitate model providers to assess the effectiveness, efficiency, and utility of the machine unlearning algorithm, and to determine whether to publish an unlearned model based on the evaluation. However, some metrics are simple and only coarsely measurable, e.g., based on accuracy, while some metrics are limited in their application, e.g., based on theory. Notably, many metrics are based on a comparison with the retrained model, e.g., metric based on similarity, which is not practical to obtain the parameters of the retrained model as a reference.

\section{Machine Unlearning Taxonomy}
\label{3}

\begin{table*}[]
\centering
\caption{Comparison of Exact and Approximate Unlearning}
\label{table:technology classification}
\scalebox{0.7}{
\renewcommand{\arraystretch}{1.5}
\begin{tabular}{|l|l|l|l|}
\hline
\textbf{Category}   & \multicolumn{1}{l|}{\textbf{Basic goal}}  & \multicolumn{1}{l|}{\textbf{Advantages}} & \textbf{Drawbacks} \\ \hline
Exact Unlearning  &  \begin{tabular}[c]{@{}l@{}}Indistinguishable distributions between \\ the unlearned and retrained model\end{tabular} & \begin{tabular}[c]{@{}l@{}}More cleaner and effectiveness unlearning \end{tabular}  & \begin{tabular}[c]{@{}l@{}}Expensive time and computational overhead, \\ Hard to implement in complex models, \\Hard to design exact unlearning algorithms \end{tabular}
\\\hline
Approximate Unlearning & \begin{tabular}[c]{@{}l@{}}Indistinguishable parameters between \\ the unlearned and retrained model\end{tabular} &\begin{tabular}[c]{@{}l@{}}Applicable to more complex models, \\ Requires minimal time investment, \\ Easier to achieve than exact unlearning\end{tabular}    & \begin{tabular}[c]{@{}l@{}}Maintains sensitive information in parameter\end{tabular}   \\ \hline     

\end{tabular}
}
\end{table*}

Training a model from scratch on $\mathcal{D}_r$ is naive and often impractical in real-world scenarios due to its high computational and time costs. 
In the centralized machine learning setting, current machine unlearning algorithms aim at addressing the issue of resource-intensive while ensuring the elimination of $\mathcal{D}_u$. These algorithms are classified into two categories based on whether they employ a retraining process: exact unlearning and approximate unlearning. The comparison of these two categories is shown in Table \ref{table:technology classification}.

\begin{itemize}
    \item Exact unlearning refers to rapidly retraining the model on an updated dataset $\mathcal{D}_r$ which $\mathcal{D}_u$ has been removed. It ensures that the distribution of the unlearned model is indistinguishable from the training-from-scratch model.
    \item Approximate unlearning avoids the need to retrain the model. As shown in Fig. \ref{fig:approximate}, the approximation can be achieved not only at the level of the model parameters (the ingrained level) but also at the level of the final activation function (the representation level) \cite{Tarun23}. Both can guarantee statistical indistinguishability with retrained models, although the guarantees of the latter prove to be relatively weaker \cite{Tarun23}.
\end{itemize}

Exact unlearning provides a theoretical guarantee for the complete elimination of $\mathcal{D}_u$ and its influence from the model through the retraining process. However, it is ineffective when dealing with complex models due to intricate mathematical computations \cite{Cao15}. 
In contrast, approximate unlearning effectively addresses the issues faced by exact unlearning. However, due to the relaxation of the need for retraining, approximate unlearning can only provide a statistical unlearning guarantee. 

\subsection{Exact Unlearning}
Due to the computational complexity required for exact unlearning,  the initial algorithm primarily targeted conventional models based on the convex function, which are more homogeneous in structure and easier to trace the influence of the data. With the great innovative efforts made by existing algorithms, they enable exact unlearning to be applicable for complex models based on the non-convex function, which can trace the data influence on interleaved neural networks, e.g., Convolutional Neural Networks (CNN), DNN, Graph Neural Networks (GNN).
The existing exact unlearning algorithms are classified into two main categories: conventional model-specific and complex model-specific, depending on whether they can be applied to non-convex function-based models.

\begin{figure*}[htbp]
\centering
\includegraphics[scale=0.5]{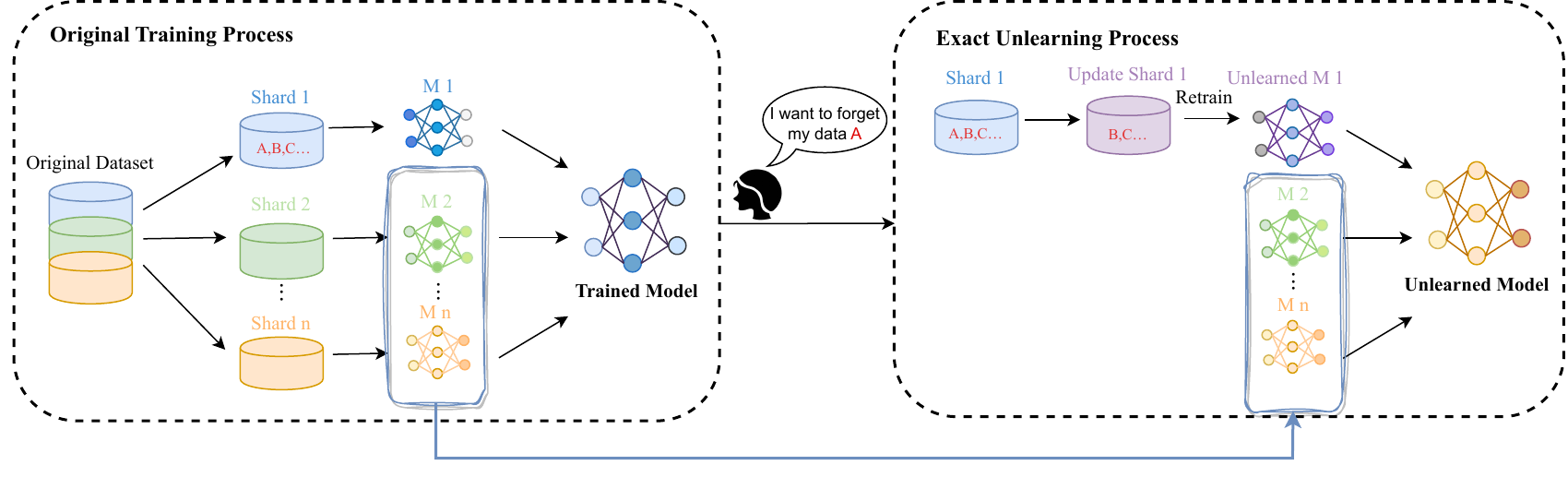}
\caption{Exact Unlearning (SISA \cite{Bourtoule21}).}
\label{fig:exact}
\end{figure*} 

\subsubsection{Conventional Model upon Convex Function} 
Conventional models are typically associated with a convex loss function, which models usually have a specific single model structure, and convex optimization makes the time consumption lower. 
Early exact unlearning algorithms are designed for such simple conventional models including Bayesian \cite{Cao15,Cao18Efficient,Jose21,Schelter21}, logistic regression \cite{Schelter20Amnesia}, support vector machines (SVMs) \cite{Brophy21,Cauwenberghs00,Tveit03,Duan07,Romero07,Karasuyama10,Chen19,Cao15,Cao18Efficient,Kashef21,Tsai14}.

Simply and directly retraining the model from scratch is ideal for achieving exact unlearning, but it becomes computationally infeasible with large datasets. 
To overcome this computational challenge, the concept of machine unlearning was first proposed by Cao \textit{et al.} \cite{Cao15} in 2015, they converted the learning algorithm into a summation form that follows the statistical query learning. When an unlearning request is received, only the corresponding small number of summations needs to be deleted from the sum, thus reducing the unlearning overhead. 
However, this method is only suitable for straightforward learning models (e.g. naive Bayes and SVMs) that can be converted into a summation form and is not adapted to deep learning. 
In 2018, Cao \textit{et al.} \cite{Cao18Efficient} introduced Karma, a causal unlearning method designed to repair damaged machine learning systems effectively. However, Karma is limited to SVMs and Bayesian-based classifiers.
Since 2020, a wealth of exact unlearning methods have been developed, each specifically tailored to different models. For example, Schelter \cite{Schelter20Amnesia} proposed an unlearning method for logistic regression that relies on decremental updates. 
Jose \textit{et al.} \cite{Jose21} designed an unlearning algorithm for PAC-Bayesian that achieves effective unlearning through information risk minimization. In addition, Kashef \cite{Kashef21} improved the unlearning efficiency for weak non-linear SVMs through decremental strategies.

\subsubsection{Complex Model with Non-Convex Function} 
Machine learning can be translated into optimization problems. Non-convex loss functions are often used in complex models. However, non-convex optimization can lead to multiple local optimal solutions, making it difficult to track data and requiring more resources than convex optimization. For example, neural network models use highly non-convex loss functions due to their multi-layer complex structure and non-linear activation. Achieving exact unlearning in this setting while minimizing time consumption is challenging.

\begin{itemize}
\item\textbf{CNN or DNN.} Ullah \textit{et al.} \cite{Ullah21} devised exact unlearning by storing the model's historical parameters, which applies to empirical risk minimization. Bourtoule \textit{et al.} \cite{Bourtoule21} introduced a notable method named SISA. The training dataset is partitioned into mutual disjoint shards, and then sub-models are trained on these shards, as shown in Fig.~\ref{fig:exact}. Upon unlearning, SISA only needs to retrain the sub-model correlated with the relevant data shard and then make the final prediction upon assembling the knowledge of each sub-model, which significantly lowers retraining computational costs. However, SISA has lower accuracy compared to a model trained on the entire unsegmented dataset and needs to maintain the whole training data. 

\item\textbf{GNN.} 
Before 2022, machine unlearning algorithms mainly focused on image and text data. Chen \textit{et al.}\cite{Chen22Graph} introduced GraphEraser based on SISA, the first exact unlearning for GNN. Unlike SISA's random data partitioning, GraphEraser offers two balanced partition methods to preserve graph structural information, this addresses the issue of applying SISA directly to graph data severely damages the graph's structure. However, GraphEraser has high graph partitioning time costs, limiting its use with the evolving graph or the multi-graph in the inductive setting. To solve this problem, Wang \textit{et al.} \cite{Wang23Inductive} introduced GUIDE, the first model-agnostic inductive graph unlearning algorithm. It ensures fairness and balance constraints in graph partitioning, outperforming GraphEraser in both time efficiency and fairness and balance scores.

\end{itemize}

\begin{table*}[]
\centering
\caption{Comparison of Approximate Unlearning Approaches}
\label{table:comparison}
\scalebox{0.6}{
\renewcommand{\arraystretch}{1.5}
\begin{tabular}{|l|l|l|l|l|l|l|}
\hline
\multicolumn{2}{|l|}{\textbf{Category}}    & \multicolumn{1}{l|}{\textbf{Advantages}}  & \multicolumn{1}{l|}{\textbf{Drawbacks}} & \textbf{Resource Cost} & \begin{tabular}[c]{@{}l@{}}\textbf {Model Utility}\end{tabular}& \begin{tabular}[c]{@{}l@{}}\textbf{Apply} \\\textbf{Model's Complex} \end{tabular} \\ \hline

\multirow{2}{*}[-2.5ex] {\begin{tabular}[c]{@{}l@{}}Data-driven \\ Approximation  \end{tabular}}
&\begin{tabular}[c]{@{}l@{}}Data Isolation \\ \cite{Gupta2021Adaptive,Neel21,He2021deepobliviate}\end{tabular} &  \begin{tabular}[c]{@{}l@{}}Applicable to most model types\end{tabular} & \begin{tabular}[c]{@{}l@{}} Unlearning is not complete \end{tabular}
& \CIRCLE\Circle\Circle\Circle
&\CIRCLE\CIRCLE\CIRCLE\Circle
&\CIRCLE\CIRCLE\CIRCLE\Circle    \\\cline{2-7}

& \begin{tabular}[c]{@{}l@{}} Data Modification \cite{Felps2021Class} \\ \cite{Graves21,Tarun21,Chundawat23Zero-Shot,Chen23} \end{tabular} & \begin{tabular}[c]{@{}l@{}}Unlearning process is easy to achieve\end{tabular} &\begin{tabular}[c]{@{}l@{}}Impacts the model's utility, \\ Consumes storage resources\end{tabular}   & \CIRCLE\CIRCLE\Circle\Circle
& \CIRCLE\CIRCLE\Circle \Circle   
& \CIRCLE\CIRCLE\CIRCLE\Circle     \\\hline

\multirow{5}{*}[-2.5ex]{\begin{tabular}[c]{@{}l@{}}Model-driven \\ Approximation \end{tabular}}
& \begin{tabular}[c]{@{}l@{}}Influence-based \cite{Guo20} \\ \cite{Izzo21,Warnecke23,Wu23GIF,Wu23Certified} \end{tabular}
& \begin{tabular}[c]{@{}l@{}}No need to store additional information\end{tabular} &\begin{tabular}[c]{@{}l@{}}Relatively complex calculation \end{tabular} 
& \CIRCLE\CIRCLE\CIRCLE\Circle 
& \CIRCLE\CIRCLE\CIRCLE\Circle   
& \CIRCLE\CIRCLE\CIRCLE\CIRCLE   \\
\cline{2-7}

&\begin{tabular}[c]{@{}l@{}}Fisher-based \\ \cite{Golatkar20Eternal,Golatkar20Forgetting,Golatkar21} \end{tabular}
& \begin{tabular}[c]{@{}l@{}}Maintain original model performance\end{tabular} &\begin{tabular}[c]{@{}l@{}}Relatively high time consumption \end{tabular}
& \CIRCLE\CIRCLE\CIRCLE\Circle  
&\CIRCLE\CIRCLE\CIRCLE\Circle   
& \CIRCLE\CIRCLE\CIRCLE\Circle   \\\cline{2-7}

&\begin{tabular}[c]{@{}l@{}}Distillation-based \\ \cite{Chundawat23Bad,Kurmanji23} \end{tabular}
& \begin{tabular}[c]{@{}l@{}}More suitable for complex models\end{tabular} &\begin{tabular}[c]{@{}l@{}}Higher computational and time cost\end{tabular}
& \CIRCLE\CIRCLE\CIRCLE\CIRCLE 
&\CIRCLE\CIRCLE\CIRCLE\Circle        & \CIRCLE\CIRCLE\CIRCLE\Circle   \\\cline{2-7}

&\begin{tabular}[c]{@{}l@{}} Gradient-based \\ \cite{Wu20,Graves21} \end{tabular}
&\begin{tabular}[c]{@{}l@{}}Less computational resource consumption \end{tabular}  &\begin{tabular}[c]{@{}l@{}} Large storage consumption,\\Unlearning is not complete \end{tabular}
& \CIRCLE\Circle\Circle\Circle  
&\CIRCLE\CIRCLE\Circle\Circle        & \CIRCLE\Circle\Circle\Circle   \\\hline

\end{tabular}
}
\end{table*}

\subsection{Approximate Unlearning}

More methods have been conducted on approximate unlearning than on exact unlearning, as the former does not require retraining. Based on different approximate unlearning strategies, strategies that focus on manipulating the data are categorized as data-driven approximation, while strategies that directly revise the original trained model are categorized as model-driven approximation. Table \ref{table:comparison} compares and analyzes various approximate unlearning categories.

\subsubsection{Data-driven Approximation}
There are two ways to manipulate data: data isolation and data modification. Data isolation can minimize the impact of the data, whereas data modification changes the model's understanding of the $\mathcal{D}_u$.
\begin{itemize}

\item\textbf{Data Isolation.}
Model providers first separate $\mathcal{D}_u$ from $\mathcal{D}_r$, i.e., split the training dataset into several disjoint partitions (sub-datasets) like SISA. Sub-models are then trained on corresponding sub-datasets, and aggregated to provide services. When unlearning is requested, the corresponding sub-dataset is identified to unlearn the corresponding sub-models unlike SISA, there is no need to retrain the sub-models.

In 2021, Neel \textit{et al.} \cite{Neel21} proposed a gradient-based removal algorithm that performs several gradient descent steps to unlearn the corresponding sub-model. The algorithm offers a trade-off between runtime and accuracy for sufficiently high-dimensional data.
Gupta \textit{et al.} \cite{Gupta2021Adaptive} introduced streaming unlearning that can handle adaptive sequences of forgetting requests with a strong provable guarantee. Specifically, they use a variant of SISA and employ DP for private aggregation. He \textit{et al.} \cite{He2021deepobliviate} proposed an unlearning algorithm that can be applied to DNN. The affected sub-models will be retrained (not fully retrained), where the retraining termination is based on the trend of residual memory tendency. Finally, they build an unlearned model by combining the retrained models and unaffected models.

\begin{figure}[htbp]
\includegraphics[scale=0.51]{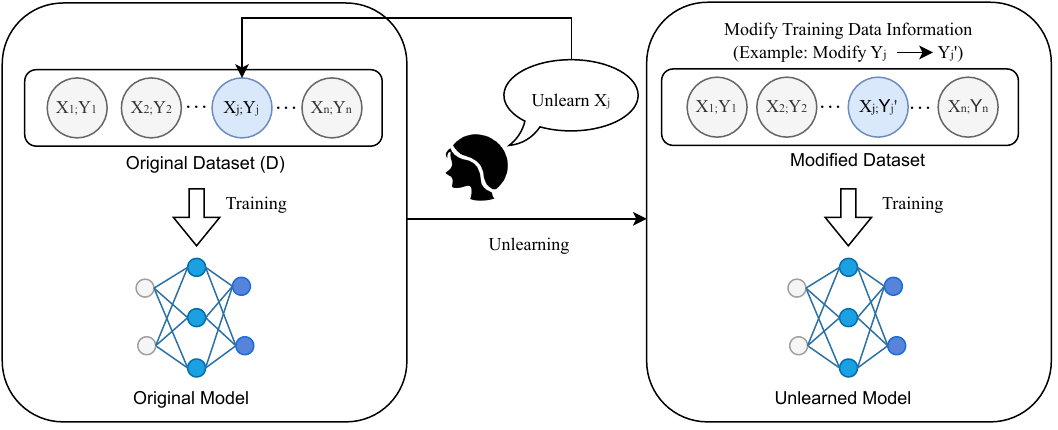}
\caption{Data Modification.}
\label{fig:approximate}
\end{figure} 

\item\textbf{Data Modification.}
As shown in Fig.~\ref{fig:approximate}, data modification refers to altering the data information in the training dataset, (e.g., modifying the labels correspond to $\mathcal{D}_u$ \cite{Graves21}), and then performing several iterations of fine-tuning on this new dataset to achieve unlearning.

Graves \textit{et al.} \cite{Graves21} proposed amnesiac unlearning, which stores parameter updates about $\mathcal{D}_u$ during training. When a forgetting request is received, the $\mathcal{D}_u$ are relabeled with random labels, and the corresponding updates are subtracted from the model parameters. 
Then unlearning is accomplished by fine-tuning the model for several epochs. 
Nonetheless, this process requires significant storage capacity to save parameter updates and may also reduce model performance. 
Felps \textit{et al.} \cite{Felps2021Class} introduced a DNN model lifecycle maintenance process that establishes how to handle specific data redaction requests.
Similar to \cite{Graves21}, unlearning is implemented by poisoning the labels of $\mathcal{D}_u$ within incremental model updates. 
Tarun \textit{et al.} \cite{Tarun21} combined the noise matrix along with the samples in the $\mathcal{D}_r$ and trained the model for one epoch to perform unlearning.
The final model demonstrates outstanding performance in unlearning the targeted data classes. 
Note that \cite{Chundawat23Zero-Shot} extended \cite{Tarun21} to the setting where no original training data is available. 
Chen \textit{et al.} \cite{Chen23} introduced boundary unlearning, which involves modifying the decision boundary of the original trained model to mimic the decision behavior of the retrained model. They provided two boundary shift methods to unlearn specific classes: boundary shrink and boundary expanding. The former disrupts the decision boundary of the forgetting class by allocating its features to other classes, while the latter disperses the activation of the forgetting class by remapping an additional class assigned to $\mathcal{D}_u$.

\end{itemize}

\subsubsection{Model-driven Approximation}
By directly manipulating the model parameters, unlearned models can be indistinguishable from retrained models in their parameter space. This is achieved through a variety of techniques, mainly relying on techniques such as Influence Function \cite{Koh17,Giordano19}, Fisher Information Matrices (FIM) \cite{Martens_2014}, Knowledge Distillation, and Stochastic Gradient Descent (SGD).

\begin{itemize}

\item {\textbf{Influence Function-based.} 
The influence function \cite{Koh17,Giordano19} is employed to evaluate the influence of $\mathcal{D}_u$ on the trained model's parameters. Then, by updating the model to negate this influence as:
\begin{equation}
{w_r} = w + {B^{ - 1}}\Delta, 
\end{equation}
where ${B^{ - 1}}$ is the second-order derivative for the loss function on the $\mathcal{D}_r$, $\Delta$ is the derivative for the loss function on the $\mathcal{D}_u$, ${B^{ - 1}}\Delta $ is the influence that $\mathcal{D}_u$ has on the model.

Guo \textit{et al.} \cite{Guo20} used Newton's method to estimate the influence of $\mathcal{D}_u$ on parameters and maximized the elimination of this influence. However, this method is only applicable to linear models.
Similarly, Izzo \textit{et al.} \cite{Izzo21} employed an influence function on the convex loss function. Their method improves the runtime efficiency of \cite{Guo20} but is challenging to fit non-convex models.
In contrast, Warnecke \textit{et al.} \cite{Warnecke23} shifted the focus of unlearning requests from samples to features and labels. They translate the effects of data into closed-form updates of model parameters---closed-form refers to a mathematical expression that provides a direct solution for the specified variables. 
These updates can be calculated directly without iteration and contribute to the correction of features and labels learned in the model. However, the effectiveness of unlearning decreases as the number of affected features and labels increases.

Traditional influence functions face challenges when applied directly to GNN due to the inherent data dependencies within graphs \cite{Wu23GIF,Wu23Certified,Li23,Said23}. To address this issue, Wu \textit{et al.} \cite{Wu23Certified} proposed the certified edge unlearning, which enables the removal of edges from the model. Similar to \cite{Warnecke23}, they carefully investigated the dependency between data and redefined the unlearning as finding a closed-form update for the model parameters, using the influence function to calculate the update effectively. 
Furthermore, Wu \textit{et al.} \cite{Wu23GIF} extended \cite{Wu23Certified} to tasks for graph node unlearning, edge unlearning, and feature unlearning. The core idea revolves around the incorporation of a loss term for the influenced neighbors alongside the conventional influence function, which allows an efficient and accurate assessment of how parameters respond to a small amount of perturbation in the dataset.
}

\item {\textbf{FIM-based.}
When using Newton's method to obtain the optimal value, the computation of the Hessian matrix can be very large. To improve unlearning efficiency, the FIM \cite{Martens_2014} can be used on $\mathcal{D}_r$ to approximate the Hessian matrix. Meanwhile, optimal noise is injected to unlearn $\mathcal{D}_u$. The unlearned model is given by:
\begin{equation}
{w_r} = w - {F^{ - 1}}{\Delta _{_R}} + b,
\end{equation}
where ${F^{ - 1}}$ is the FIM on the $D_{r}$, ${\Delta _{_R}}$ is the gradient about the loss function on the $D_{r}$, $w - {F^{ - 1}}{\Delta _{_R}}$ corresponds to the corrective Newton step, $b$ corresponds to the optimal noise added \cite{Wang22FU}.

Golatkar \textit{et al.} \cite{Golatkar20Eternal} proposed a robust unlearning algorithm based on the noisy Newton update, which can erase information about a specific class while replacing the Hessian matrix with FIM to improve efficiency. \cite{Golatkar20Forgetting} generalized \cite{Golatkar20Eternal} to different objective functions. However, the scalability of \cite{Golatkar20Eternal,Golatkar20Forgetting} decreases with a size increase in the training dataset. Because the computation of the unlearning step exhibits quadratic growth with dataset size.
Golatkar \textit{et al.} \cite{Golatkar21} solved this problem and introduced an effective unlearning method for mixed-privacy scenarios. This method partitions the training data into core and user data. Core data consists of general information utilized for pre-training and is crucial to retain. User data typically includes the information that users want to be removed. 
The core weights are learned using a non-convex algorithm, while the user weights are obtained through strongly convex quadratic optimization. Simply setting the user weights to zero removes the influence of their data. However, as the dataset needs to remain static during the pre-training phase, it may not be suitable for many practical applications. 

}

\item {\textbf{Knowledge Distillation-based.} 
Knowledge distillation enables the training of a student model to selectively mimic the knowledge of a larger teacher model, allowing for filtering out sensitive information about $\mathcal{D}_u$ while maintaining the utility of the student model.

Chundawat \textit{et al.} \cite{Chundawat23Zero-Shot} employed a band-pass filter to block the flow of sensitive information from the teacher to the student model. However, this method may not be suitable for large-scale models. In a subsequent study \cite{Chundawat23Bad}, they revised \cite{Chundawat23Zero-Shot} by using a pair of (competent/incompetent) teachers to manage the student. The misinformation about $D_{u}$ from the incompetent teacher is transferred to the student, helping to unlearn samples. While this method is highly efficient, it damages model utility \cite{Kurmanji23}. 
Kurmanji \textit{et al.} \cite{Kurmanji23} introduced an application-dependent unlearning method based on the teacher-student formulation, which is adaptable to diverse applications. Specifically, the original model is designated as the teacher, and the unlearned model functions as the student. The student model selectively disobeys an all-knowing teacher, to inherit only knowledge unrelated to $\mathcal{D}_u$.
}

\item{\textbf{Gradient-based.}
Gradient-based unlearning approximates the retrained model by correcting the SGD steps.

Wu \textit{et al.}\cite{Graves21} leveraged gradient descent to trace and exploit the data's provenance to achieve rapid incremental model updates. However, this method is only applicable to regression models. Additionally, they proposed DeltaGrad \cite{Wu20}, which utilizes Quasi-Newton methods to eliminate gradients associated with $\mathcal{D}_u$ based on cached intermediate parameters. Nevertheless, it is not well-suited for unlearning a substantial amount of data. 
}
\end{itemize}

\subsection{Debate on Approximate Unlearning}
Recently, some researchers have argued that defining approximate unlearning as generating unlearned models that are indistinguishable from retrained models in parameters space is ill-considered. 
Firstly, Thudi \textit{et al.} \cite{Thudi22On} posited theoretical proof that one can obtain the same model without alterations from training on a pair of non-overlapping datasets. This implies that attaining a specific location in the parameters' universe is not a sufficient condition for unlearning, essentially questioning the defination of approximate unlearning. Moreover, Tarun \textit{et al.} \cite{Tarun21} argued that utilizing the retrained model parameters as a comparative benchmark for approximate unlearning quality is unreliable, due to the potential existence of numerous parametric configurations capable of efficient unlearning (the retrained model's parameter is only one of them). Even if there is a significant difference in parameters between unlearned and retrained models, it does not necessarily mean that the unlearning process has failed \cite{Kurmanji23,Yang20,Goel22}.
Therefore, researchers try to achieve approximate unlearning without aiming for indistinguishability, Wang \textit{et al.} \cite{Wang23KGA} avoided forcing model parameters to conform to a particular distribution. Instead, it preserves the differences in distribution between unlearned and retrained models, thereby making it adaptable for more applications (e.g., natural language processing). Lin \textit{et al.} \cite{Lin23Knowledge} attempted to define machine unlearning from a knowledge perspective, proposing a method based on knowledge transfer for knowledge-level machine unlearning.

\vspace{0.15cm}
\textbf{Summary:} Table \ref{table:approximate} summarizes main machine unlearning algorithms. Exact unlearning proves effective in ensuring comprehensive unlearning while preventing any attempts by adversaries to extract valuable information from the unlearned model. However, its implementation in time or computation is relatively intensive, particularly with DL models. Additionally, the unlearning request occurs repeatedly, rather than being a one-time occurrence, which further aggravates resource consumption. Therefore, great efforts have been made to devise efficient exact unlearning algorithms, aiming to reduce runtime and computation to a level far shorter than the naive retrain from scratch. 
The prevalence of approximate unlearning is motivated by more computationally efficient unlearning, although with a trade-off of the unlearning degree and sensitive information residual in the model. The primary goal of approximate unlearning is to develop algorithms that guarantee complete unlearning.

\begin{table*}[]
\centering
\caption{Summary of Studies in Approximate Unlearning}
\label{table:approximate}
\scalebox{0.75}{
\renewcommand{\arraystretch}{1.5}
\begin{tabular}{|l|l|l|l|l|l|l|}
\hline
\multicolumn{3}{|l|}{\textbf{Category}}              & \multicolumn{1}{l|}{\textbf{Papers}} & \multicolumn{1}{l|}{\textbf{Year}} & \multicolumn{1}{l|}{\textbf{Applicable Model}}  & \multicolumn{1}{l|}{\textbf{Type of   Unlearn Request}}  \\ \hline

\multirow{12}{*}{\rotatebox{90}{\textbf{Exact Unlearning}}} &\multicolumn{2}{l|}{\multirow{5}{*}{\makecell[c]{Conventional Model upon Convex Function}}}     & \cite{Cao15}          & 2015 &  \begin{tabular}[l]{@{}l@{}}  Straightforward Learning Models \\ (e.g. naive Bayes) \end{tabular}   & Samples \\\cline{4-7}

 &\multicolumn{2}{l|}{}& \cite{Cao18Efficient}    & 2018         & SVMs and Bayesian-based Classifiers & Subset                   \\\cline{4-7}
 
& \multicolumn{2}{l|}{}&  \cite{Schelter20Amnesia}    & 2020       & Logistic Regression                & Samples         \\\cline{4-7}



& \multicolumn{2}{l|}{}&\cite{Jose21}      & 2021         & Bayesian   & Subset   \\\cline{4-7}
 
& \multicolumn{2}{l|}{}&\cite{Kashef21}    & 2021        & SVMs  & Samples  \\\cline{2-7}

&\multicolumn{2}{l|}{\multirow{5}{*}{\makecell[l]{Complex Model with Non-Convex Function }}}        & \cite{Ullah21}     & 2021 & Models with Non-convex Functions   & Samples \\\cline{4-7}
& \multicolumn{2}{l|}{}& \cite{Bourtoule21} & 2021  & DNN    & Batches, Sequences  \\\cline{4-7}

 &\multicolumn{2}{l|}{}& \cite{Yan22ARCANE}       & 2022       & DNN  & Samples   \\\cline{4-7}
 
&\multicolumn{2}{l|}{} &\cite{Chen22Graph}   & 2022 &  GNN     & Nodes, Edges  \\\cline{4-7}

&\multicolumn{2}{l|}{} &\cite{Wang23Inductive}      & 2023         & GNN  & Nodes, Edges    \\ \hline

\multirow{21}{*}{\rotatebox{90}{\textbf{Approximate Unlearning}}}&\multirow{8}{*}{\makecell[l]{Data-driven Approximation}}  
&\multirow{3}{*}{Data Isolation} & \cite{Gupta2021Adaptive}         & 2020 &  SGD-Based Models     & Samples  \\\cline{4-7}

&&  & \cite{Neel21}      & 2021 & Convex Models      & Non-adaptive Sequences   \\\cline{4-7}

&&  &\cite{He2021deepobliviate}          & 2021 &  DNN     & Samples  \\\cline{3-7}

&&\multirow{5}{*}{Data Modification}  &\cite{Graves21}    & 2021 &  DNN  & Classes, Samples \\\cline{4-7}

&&  &\cite{Felps2021Class}    & 2021 &  DNN  & Batches \\\cline{4-7}

&&  &\cite{Tarun21}     & 2021 &  CNN  & Samples \\\cline{4-7}


&&  &\cite{Chen23}    &2023  & DNN   & Classes \\\cline{2-7}
 
&\multirow{13}{*}{\makecell[l]{Model-driven Approximation}}     
& \multirow{5}{*}{Influence-based}  &\cite{Guo20}        & 2020 &  Linear Models& Samples  \\\cline{4-7}

&&  &\cite{Izzo21}       & 2021 &  \begin{tabular}[c]{@{}l@{}}Logistic and Linear Regression Models\end{tabular}& Batches\\\cline{4-7}

& & & \cite{Warnecke23}   & 2023 &  Convex or Non-convex Models& Features, Labels \\\cline{4-7}

& & &\cite{Wu23GIF}        & 2023 &  GNN & Nodes, Edges, Features   \\\cline{4-7}

& & & \cite{Wu23Certified}        & 2023 & GNN& Edges  \\\cline{3-7}

& &\multirow{3}{*}{Fisher-based} &  \cite{Golatkar20Eternal}   & 2020 &  DNN & Particular set of training data\\\cline{4-7}

& & & \cite{Golatkar20Forgetting}   & 2020 &  DNN & Samples\\\cline{4-7}

&&  & \cite{Golatkar21}   & 2021 &  DNN & Samples  \\\cline{3-7}

&& \multirow{3}{*}{Distillation-based} & \cite{Chundawat23Zero-Shot}  & 2023 &  DNN   & Samples, Classes \\\cline{4-7}
 
&&  & \cite{Chundawat23Bad}  & 2023&  DNN & Classes\\\cline{4-7}

&&  & \cite{Kurmanji23}   & 2023 &  DNN & Subset, Classes \\\cline{3-7}

&& \multirow{2}{*}{Gradient-based} & \cite{Graves21}         & 2020 &  Regression Models  & Subset\\\cline{4-7}

& & & \cite{Wu20}         & 2020 &  SGD-Based Models     & Samples  \\\hline

\end{tabular}
}
\end{table*}

\section{Distributed Unlearning}
\label{4}
Under the dual pressure of big data and complex models, large-scale ML training in Centralized Machine Learning (CML) poses significant challenges in both computational power and storage capacity. To address these challenges, Distributed Machine Learning (DML) leverages multiple computational nodes in parallel to train ML models. Notably, DML can greatly mitigate privacy information leakage of the user's local data since these data cannot be accessed by others. There are different DML schemes including Federated Learning (FL), Split Learning, Peer-to-peer, and Private Aggregation of Teacher Ensembles (PATE). 

There still exists privacy leakage risks in DML. For example, FL is vulnerable to exposure of user preferences through Preference Profiling Attacks (PPA) \cite{Zhou23}. Moreover, in many contexts, such as when the data is poisoned or invalid, it is necessary to delete the data locally, while unlearning the impact of the data on the global model. Therefore, it is necessary to deploy machine unlearning in a DML setting, referred to as \textit{Distributed Unlearning} (DU).

\subsection{Challenges}
However, most existing MU algorithms are designed under CML and do not apply to DML, primarily owing to the disparate ways they gain information from data. There exist three principal challenges---since current DU algorithms focus on FL settings, we analyze the challenges facing DML mainly taking FL as an example: \par
\begin{itemize}
\item{\textbf{Data Availability Perspective.}} From the perspective of data availability, three primary issues arise. First, data contributors (clients) are not sharing their data due to privacy concerns, the server does not have access to the local training datasets (e.g., FL). Therefore, server-side retraining becomes virtually impossible, only approximate unlearning can be performed on the server. Second, the frequent drop-in and drop-out clients pose a significant challenge for servers attempting to recall former clients for unlearning operations, let alone retraining them from scratch \cite{Zhang23FedRecovery}. Third, clients such as edge devices may discard data after local training due to limited storage capacity, so the client may no longer have the same dataset used during the training phase \cite{Wu22Federated}.
\item{\textbf{Model Parameters Perspective.}} From the perspective of model parameters, several issues need to be considered. 
Firstly, training in DML is interactive, e.g., clients continuously share the knowledge learned from the local training dataset with other clients through the global model \cite{Liu22right}. 
Secondly, updating model parameters in DML is a complex entangled incremental process, e.g., in FL, each local update of each client is based on the updates of all previous clients \cite{Liu22right}. 
Thirdly, DML involves significantly more stochasticity than CML \cite{Wu22Federated}, e.g., the clients involved in each training epoch are randomly selected in FL, and the inherent randomness in each client's local training process. 
Any minor perturbations due to stochasticity could potentially render a domino effect throughout the subsequent training process \cite{Wu22Federated}. The combination of these three issues severely complicates the traceability of data based on global model parameters, compromising the central server's ability to selectively unlearn data from a specific client.

\item{\textbf{Resource Overhead Perspective.}} From the perspective of resource overhead, DML usually has a high communication and time overhead compared to CML due to the necessity for the server to exchange information (e.g., local or global model parameters) with clients. Therefore, ideal machine unlearning should not bring additional expensive communication and time consumption to DML.
\end{itemize}

To achieve both efficiency and effectiveness, DU must overcome not just the hurdle faced by MU in CML, but also the unique three above-mentioned challenges \cite{Wang23BFU}. Therefore, unlearning in distributed learning is much less explored than that in centralized learning \cite{Liu22right}.

\begin{figure*}[]
    \centering 
    \centering
    \subfloat[]{\includegraphics[width=2.5in]{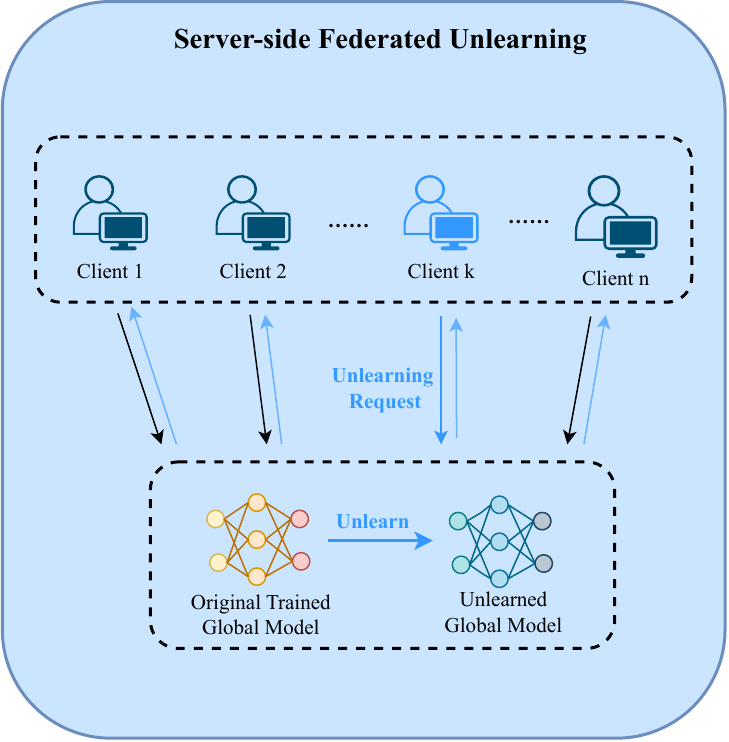}%
\label{FU}}
\hfil
\subfloat[]{\includegraphics[width=2.5in]{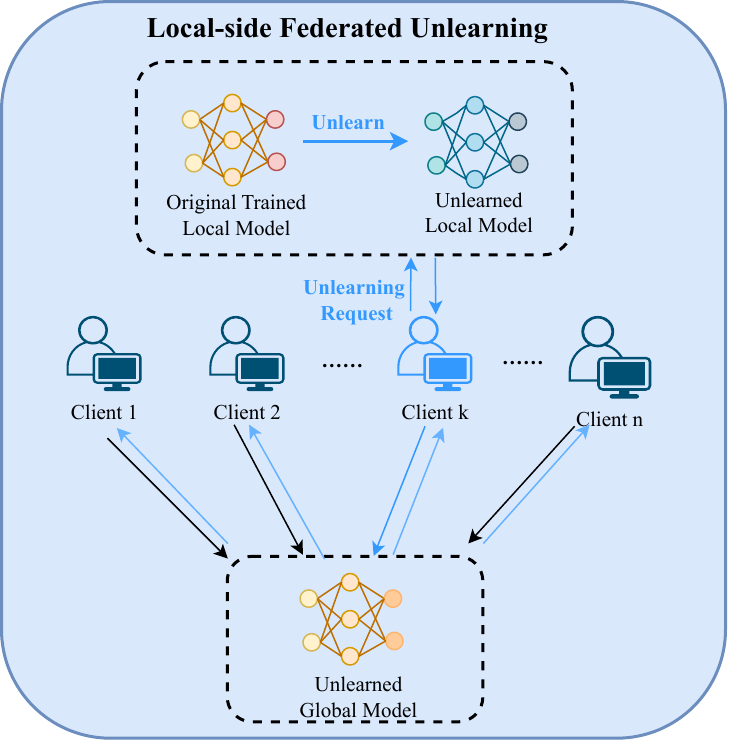}%
\label{FU2}}
\caption{Overview of Federated Unlearning.}
\label{F}
\end{figure*}

\subsection{Federated Unlearning}
Distributed unlearning primarily focuses on FL, with very little attention given to split learning \cite{Yu23Split}, while other distributed schemes, e.g., peer-to-peer remain unexplored so far. Therefore, we focus on distributed unlearning under federated learning settings.

Federated learning \cite{Fu22,Zhou20} is conceived as a methodology to safeguard user privacy. During the model training process, clients upload only their local model parameters, avoiding the need to share sensitive local raw training data with the server. Although FL does not directly use training data from clients, it indirectly involves the analysis of data generated locally via uploaded local parameters. The deployment of machine unlearning within the FL, often referred to as \textit{Federated Unlearning} (FU), can significantly enhance the security and robustness of the model through unlearning sensitive data \cite{Wang23BFU}. 

FU algorithms to date can be divided into two categories, depending on which entity mainly performs the unlearning: Server-side Federated Unlearning and Local-side Federated Unlearning.
\subsubsection{Server-side Federated Unlearning} As local clients never share raw training data with the server during collaborative training, the only way to implement unlearning at the server is to use approximate unlearning on the global model. Therefore, no computational and communication action is required from clients. Fig.~\ref{FU} depicts the overview of server-side federated unlearning.

Liu \textit{et al.} \cite{Liu21FedEraser} pioneered the study of client-level unlearning, proposing FedEraser to eliminate the contribution of a client on the global model. Specifically, the server maintains all records of historical updates for each client at every FL round. These records are then refined by several rounds of a calibration training process without the forgotten client, accelerating the unlearning process. Nonetheless, this method offers limited improvements compared to retraining from scratch. This is primarily because clients are still required to train the local model to rectify their historical updates, which renders additional rounds of communication between clients and the server.
Regarding the issue of FedEraser, Wu \textit{et al.} \cite{Wu22Federated} emphasized the need to decrease the number of iterations between server and client interactions due to the substantial time and energy consumed by communication, particularly in DNN. 
Based on this premise, they proposed a solution that directly subtracts the accumulated historical updates from the federated global model parameters and utilizes the knowledge distillation to maintain the model's performance, which effectively eliminates a client's contribution. 
However, their method requires the server to possess additional outsourced unlabeled data, which may not be feasible in certain high-privacy applications (e.g., medical systems).
Zhang \textit{et al.} \cite{Zhang23FedRecovery} applied DP to FU, leveraging clients' historical submissions to eliminate a weighted sum of gradient residuals from the global model. Further, they structured the Gaussian noise such that the unlearned and the retrained model became statistically indistinguishable, effectively removing the influences of individual clients on the global model.

Note that \cite{Wu22Federated,Zhang23FedRecovery}, and FedEraser have considerable limitations as they focus only on client-level unlearning, thus limiting their practical use when the need is to erase only a subset of the training data e.g., a specific class. They also require storing historical updates, which creates significant memory overhead, especially for state-of-the-art large models. 
To address these issues, 
Wang \textit{et al.} \cite{Wang22FU} exploited the CNN channel pruning method to guide the category-level FU. They employed Term Frequency Inverse Document Frequency (TF-IDF) \cite{Paik13} to quantify the correlation between channels and categories. The channels with high TF-IDF scores play a more substantial role in distinguishing the forgotten categories and, thus need to be pruned to erase their contributions to the global model. 
The performance of unlearned models is subsequently restored by fine-tuning based on the remaining dataset.
To serve a broader range of unlearning request needs, more generalized unlearning algorithms have been proposed. Wu \textit{et al.} \cite{Wu22Guarantee} provided the first comprehensive investigation into a general pipeline capable of handling three common types of FU requests: class unlearning, client unlearning, and sample unlearning. They reconsider how the training data impacts the global model performance and achieve unlearning through the integration of reverse stochastic gradient ascent and elastic weight consolidation.

\subsubsection{Local-side Federated Unlearning} Since only approximate unlearning can be performed on the server side, this leads to the preservation of residual sensitive information related to $\mathcal{D}_u$ in the global model. Consequently, some studies suggest that the best unlearning mechanism in FL is to perform retraining among clients \cite{Liu22right,Wang23BFU,Che23}. 
In this case, the unlearning operation must be performed on the local side. However, since the computational resources of edge devices are typically restricted, the primary concern currently is to design a fast unlearning method on the local side that is low-cost and maintains the utility of the global model. Fig. \ref{FU2} depicts the overview of local-side federated unlearning.

Liu \textit{et al.} \cite{Liu22right} developed a rapid retraining method to fully erase data points from a trained global model. This method employed a diagonal empirical FIM to approximate the Hessian for Quasi-Newton optimization, achieving a low cost while preserving model utility via the momentum technique. Nevertheless, their method is only applicable to models with convex loss functions.
Wang \textit{et al.} \cite{Wang23BFU}~proposed an algorithm with parameter self-sharing based on variational Bayesian inference to unlearn specific data points. It can mitigate accuracy degradation caused by unlearning and balance the trade-off between the unlearning effectiveness and model utility. 
Che \textit{et al.} \cite{Che23} introduced the first method for simultaneously implementing the training and unlearning process in FL, based on their previous work in centralized learning \cite{Zhang22Prompt}. By utilizing \cite{Zhang22Prompt}, a local unlearned model is trained on each local client. Then, leveraging the theory of nonlinear functional analysis to refine the local unlearned model as output functions of a Nemytskii operator, ensuring that the global unlearned model closely parallels the performance of each local unlearned model and significantly speeds up unlearning.
Zhu \textit{et al.} \cite{Zhu23} provided a unique heterogeneous knowledge graph embedding unlearning derived from cognitive neuroscience. By combining retroactive interference with passive decay, it erases specific knowledge from local clients and propagates to the global model through knowledge distillation.

\begin{table*}[]
\centering
\caption{Summary of Federated Unlearning.}
\scalebox{0.8}{
\label{table:federated}
\renewcommand{\arraystretch}{1.5}
\begin{tabular}{|l|l|l|l|}
\hline
\textbf{Category}    & \multicolumn{1}{l|}{\textbf{Advantages}} & \multicolumn{1}{l}{\textbf{Drawbacks}}  & \multicolumn{1}{|l|}{\textbf{Application Scenario}} \\ \hline
Server-side  Federated Unlearning&  \begin{tabular}[c]{@{}l@{}}Less time consuming\end{tabular} 
& \begin{tabular}[c]{@{}l@{}}Incompletely   unlearning, \\Consumes storage space \end{tabular}  
& Large scale company  \\\hline
Local-side  Federated Unlearning & \begin{tabular}[c]{@{}l@{}}More complete unlearning, \\More flexible unlearning request type\end{tabular} &\begin{tabular}[c]{@{}l@{}}Limited computational ability, \\Difficult to design effective algorithm\end{tabular}    & Internet of Things   \\ \hline     
\end{tabular}}
\end{table*}

\vspace{0.15cm}
\noindent\textbf{Summary:} As shown in Table \ref{table:federated}, FU conducted on the server side can relatively promptly accomplish the unlearning request without incurring additional communication to the client. However, there are notable shortcomings. Firstly, the unlearning is not complete, with sensitive information remaining in the global model \cite{Liu22right,Wang23BFU,Che23}. Secondly, many current methods are based on information (e.g., historical updates) stored in the server during the training phase \cite{Liu21FedEraser,Wu22Federated}, while this results in a significant additional storage burden, especially for complex models. 

Federated unlearning performed on the local side can resolve shortcomings faced by the server side. It enables more complete unlearning through retraining and prevents the negative impact of the server (model provider) dishonesty, as the unlearning process occurs locally. 
However, there are also disadvantages, one is that the computational capability of edge devices is insufficient, especially when faced with a large dataset or complex model; the other is that effective rapid retraining algorithms are still challenging to achieve.

\section{Application of Machine Unlearning}
\label{6}
Machine unlearning is primarily employed to safeguard user data privacy in compliance with legal and individual requirements. In recent years, as shown in Table~\ref{table:applications}, its applicability has expanded to other applications. To begin with, it can be harnessed to optimize models and mitigate the potential harm caused by malicious, outdated, or adversarial data. This is especially pertinent in fields such as Recommendation Systems (RES) \cite{Chen22}, the Internet of Things (IoT) \cite{Fan23}, and Large Language Models (LLMs) \cite{yao2023large}. Furthermore, it serves as an effective defense mechanism to enhance the model's robustness. It can be used as the passive defense to reduce damage from data poisoning attacks and backdoor attacks as well as the active defense to make various privacy attacks (e.g., membership inference attack, property inference attack, model inversion attack) fail.

\subsection{Optimization of the Model}
Machine unlearning can optimize models in a variety of real-world scenarios, eliminating the risk of privacy leakage and the negative effects of harmful data, and thus increasing the robustness of the model. Currently, there are three main application scenarios: LLMs, RES, and IoT.

\subsubsection{Unlearning for Large Language Models} 
State-of-the-art LLMs are trained on massive internet corpora to obtain a wide range of world knowledge \cite{yao2023large}. As a representative, ChatGPT is capable of tasks such as translation and question answering \cite{Else23,Walker23}. However, the training process can make LLMs memorize and reproduce private or harmful data. This can lead to the exhibition of undesirable problems related to racism, sexism, and religious bias, which raises both legal and ethical concerns. 
In this context, machine unlearning can assist LLMs in ensuring security, adhering to ethical standards, and eliminating bias by unlearning specific data.

\textbf{Challenge.} Traditional machine unlearning methods may not be suitable for LLMs for two main reasons. First, the parameter space of LLMs is extremely large, making it difficult to track the impact of data points and the model retraining computation is extremely large. Second, while traditional unlearning methods are primarily designed for classification tasks, LLMs are knowledge-intensive and used for generative tasks \cite{Si_Zhang_Chang_Zhang_Qu_Zhang_2023}.

\textbf{Methods.} Yao \textit{et al.} \cite{yao2023large} were the first to formulate the settings, goals, and evaluations in LLMs unlearning. Eldan \textit{et al.} \cite{Eldan2023Harry} proposed an unlearning method to remove a subset of the training data from an LLMs. Firstly, they used a reinforced model to identify the tokens most related to the unlearning target. Secondly, they leveraged the model's predictions to build alternative labels for each token. Thirdly, the model was fine-tuned on these alternative labels, effectively unlearning the original text from the model's memory. Chen \textit{et al.} \cite{chen2023unlearn} introduced additional unlearning layers learned with a selective teacher-student objective into the transformers, which can identify knowledge that needs to be forgotten. A sequence of unlearning operations can be handled through the offline fusion of different trained unlearning layers. Maini \textit{et al.}  \cite{Maini_tofu} proposed tofu, a fictional unlearning task, as a benchmark to help deepen understanding of unlearning. They also provide a set of baseline results from existing unlearning algorithms (e.g. gradient difference \cite{liu2022continual}). 
The application of machine unlearning techniques has substantially enhanced the ethical sensitivity and privacy of LLMs.

\begin{table*}[]
\centering
\caption{Summary of Applications}
\label{table:applications}
\scalebox{0.65}{
\renewcommand{\arraystretch}{1.5}
\begin{tabular}{|l|l|l|l|}
\hline 
\multicolumn{2}{|l|}{\textbf{Application Scenarios}}  & \multicolumn{1}{l}{\textbf{Challenges}}   & \multicolumn{1}{|l|}{\textbf{Aim}} \\
\hline

\multirow{3}{*}[-4.5ex]{Optimization of the Model}
&\begin{tabular}[c]{@{}l@{}}Large Language Models \\ \cite{yao2023large, Eldan2023Harry,chen2023unlearn,Maini_tofu} \end{tabular}    &The parameter space of LLMs is extremely large      &Unlearning private or harmful data in LLMs 
\\
\cline{2-4}
&\begin{tabular}[c]{@{}l@{}}Recommender Systems \\ \cite{Chen22,Yuan23,Leysen23,Zhang23Recommendation,Xu23,Liu23,Christian22}\end{tabular}    &Need to consider collaborative information       &Unlearning specific data and personal preferences   
\\
\cline{2-4}
&\begin{tabular}[c]{@{}l@{}}Internet of Things \\ \cite{Fan23, Zeng23} \end{tabular}     &   The unlearning process must be fast enough &Rapid completion of model updates  \\
\hline

\multirow{2}{*}[-2.5ex]{Defense against Various Attacks}&\begin{tabular}[c]{@{}l@{}}Passive Defense \\ \cite{Chen22Graph,Backdoor,Zeng22,Wei23,Zhang23Poison} \end{tabular} &  Unable to prevent attacks in advance &Purifies compromised memories in the model\\
\cline{2-4}

&\begin{tabular}[c]{@{}l@{}}Active Defense\\ \cite{Golatkar20Forgetting,Golatkar21, Graves21,Chundawat23Zero-Shot,Stock23, Ganju18} \end{tabular} &Unaware of the target data of the adversary        &\begin{tabular}[c]{@{}l@{}}Preliminary elimination of sensitive information \end{tabular}    \\\hline
\end{tabular}}
\end{table*}

\subsubsection{Unlearning for Recommender Systems} Recommender systems are personalized information filters that analyze users' preferences from collected data and recommend the most relevant items. Through the training phase, the recommender model's parameters can memorize user behaviors, which poses a risk of privacy leakage. This led to the development of recommendation unlearning, enabling the unlearning of specific data and personal preferences in the model \cite{Liu23}.

\textbf{Challenge.} Unlearning methods designed for generic machine learning models engaged in classification tasks are not directly applicable to recommender systems. The reason is that the fundamental principle of most recommender systems is collaborative filtering, but existing MU methods ignore the collaborative information between users and items \cite{Li22, Li23}.

\textbf{Methods.} Chen \textit{et al.} \cite{Chen22} utilized the similarity of data to divide training data into balanced groups, based on which they proposed an unlearning algorithm tailored to recommender systems retain the collaborative information within the data while augmenting usability, security, and applicability of the model. 
Yuan \textit{et al.} \cite{Yuan23} turned their attention to federated recommendation systems, inspired by the log-based rollback mechanism of transactions, they have developed a method that can efficiently retract a user's contribution to the federated training process, thereby enhancing the robustness of the model and strengthening resistance against potential attacks from malicious clients. 
Furthermore, \cite{Leysen23} developed a comprehensive framework for evaluating recommendation unlearning, focusing on the verifiability, efficiency, and accuracy of such methods. Since 2023, increasing studies have designed recommendation unlearning methods based on a variety of techniques, including those based on the Influence Function \cite{Zhang23Recommendation}, Matrix Completion \cite{Xu23}, Interaction and Mapping Matrices Correction \cite{Liu23}, and even adversarial training \cite{Christian22}. Machine unlearning promotes the development of more private, secure, reliable, useful, and responsible recommender systems.

\subsubsection{Unlearning for Internet of Things} The Internet of Things is a network of interconnected physical devices and objects that can collect and exchange data over the Internet, enabling remote monitoring and control of various applications. IoT service providers frequently face the task of updating deep learning-based detection models for traffic anomaly detection due to issues like mislabeled samples, device firmware upgrades, or data contamination during service delivery. This demonstrates the urgency of applying machine unlearning.

\textbf{Challenge.} IoT is usually time-sensitive and requires real-time or near real-time data feedback, so the unlearning process for IoT must be responded and completed promptly \cite{Fan23}.

\textbf{Methods.} Fan \textit{et al.} \cite{Fan23} introduced a method called ViFLa, which groups training datasets according to the calculated unlearning probability, and then each group is regarded as an individual virtual client. ViFLa can improve the effectiveness and completeness of model updates in IoT traffic anomaly detection. Zeng \textit{et al.} \cite{Zeng23} proposed CADDEraser, an effective unlearning framework for Quality-of-Service prediction in personalized IoT, enhancing model utility after unlearning requests. Machine unlearning can substantially enhance the security, availability, fidelity, and privacy dimensions within IoT systems.

\begin{figure*}
\centering
\includegraphics[scale=0.5]{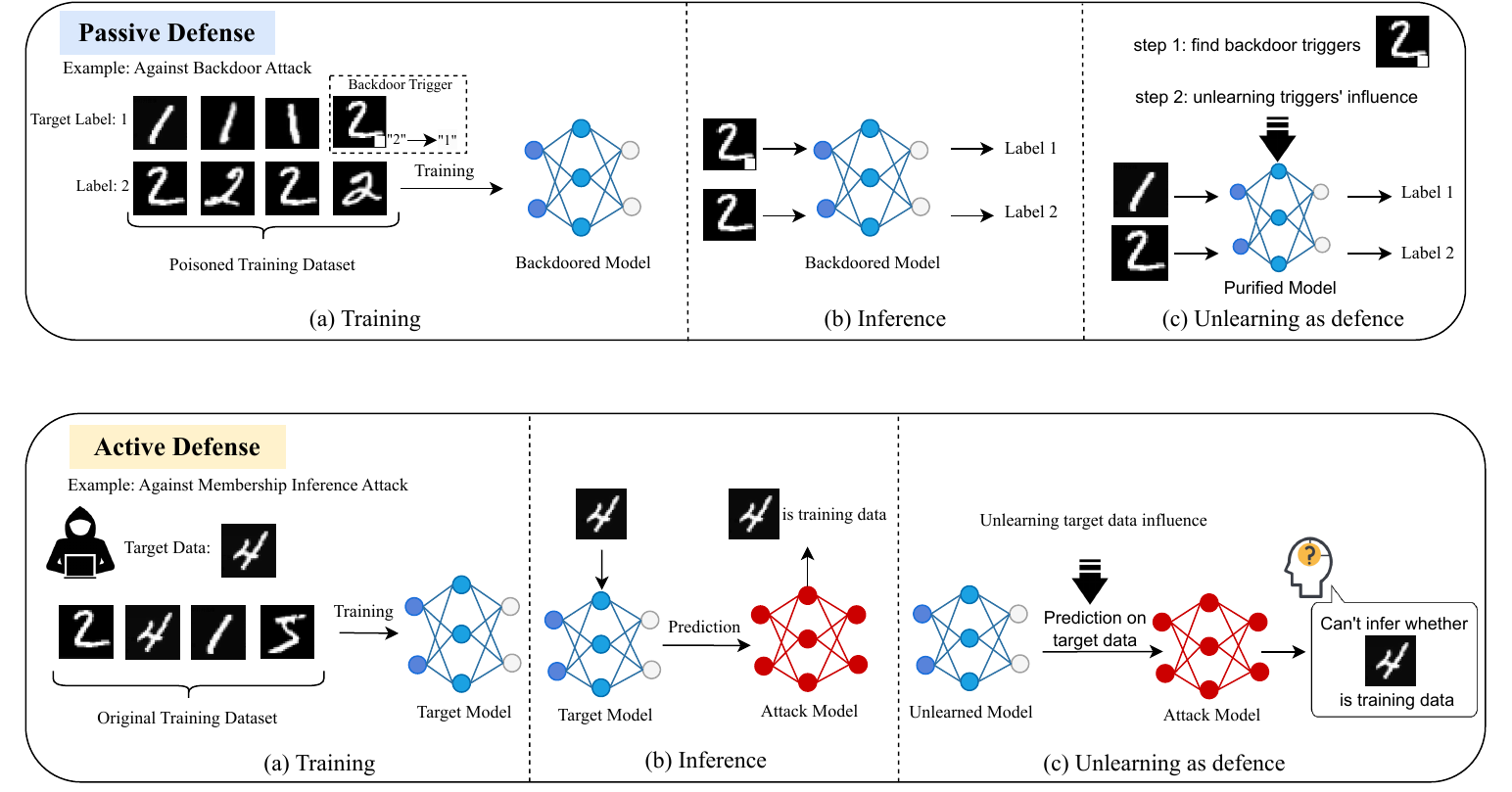}
\caption{Defence Strategies.}
\label{fig:defence}
\end{figure*}

\subsection{Defense against Attacks}
As shown in Fig.~\ref{fig:defence}, when confronted with data poisoning or backdoor attacks, machine unlearning can serve as a passive defense to clean up the negative impact of attacks on the model and restore model utility. Moreover, in the face of various privacy attacks, machine unlearning can serve as an active defense to unlearn sensitive privacy data in advance, preventing adversaries from inferring private information related to that data.

\subsubsection{Passive Defense}
After the model suffers from data poisoning or backdoor attacks~\cite{gao2020backdoor}, machine unlearning purifies compromised memories that stem from a malicious adversary by unlearning poisoned data or backdoor triggers.

\begin{itemize}
\item {\textbf{Defense of Data Poisoning Attack.}}
Data poisoning attacks refer to an adversary strategically inserting a handful of meticulously crafted poisoned samples into a model's training dataset. During the training or fine-tuning process, these samples poison the model. As a result, the model exhibits anomalous behavior during the testing phase. For instance, benign samples may be misclassified as malicious, while genuine malicious data bypass detection, compromising the integrity and usability of the model.

In \cite{Chen22Graph}, an adversary crafts malicious samples into Zozzle's training data by injecting features not found in any benign samples. The defense process is as follows: Initially, feature extraction is performed. Secondly, if the chi-value of a feature does not meet the threshold, that feature is targeted for an unlearning process and is consequently forgotten from the model. Results indicate that this defense mechanism was highly successful, almost as if the data poisoning attack had never occurred.

\item {\textbf{Defense of Backdoor Attack.}}
Backdoor attacks refer to an adversary injecting a backdoor into the model during the training process, enabling remote access and control. When this backdoor is not triggered, the attacked model behaves similarly to a regular model~\cite{peng2024model}. However, once the hidden backdoor is activated, the attacked model then engages in specific behaviors \cite{Liu22Backdoor}.

In 2022, Zeng \textit{et al.} \cite{Zeng22} proposed universal adversarial perturbation to remove backdoors. 
However, this method assumes that the backdoor can be activated by the same trigger regardless of the sample it is embedded in. 
This means it lacks defense against more advanced attacks using sample-specific or non-additive triggers \cite{Wei23}. Liu \textit{et al.} \cite{Liu22Backdoor} proposed BAERASER, first using an entropy maximization-based generative model for trigger pattern recovery, extracting the trigger patterns infected by the victim model. It then employs these recovered patterns to reverse the backdoor injection procedure and prompts the model to erase polluted memories using a specifically designed gradient ascent-based machine unlearning method, the results show that the backdoor effects can be effectively removed. 
In 2023, Zhang \textit{et al.} \cite{Zhang23Poison} proposed an adversarial unlearning method along with label smoothing to address the backdoor removal issue from the trained beam selection model. 
Wei \textit{et al.} \cite{Wei23} proposed shared adversarial unlearning, the first step of which is to generate shared adversarial examples (SAEs), and then, unlearn the SAEs so they could be correctly classified by the purified model, mitigating the backdoor effect in the purified model.

\begin{figure*}[htbp]
    \centering
    \includegraphics[width=0.8\linewidth]{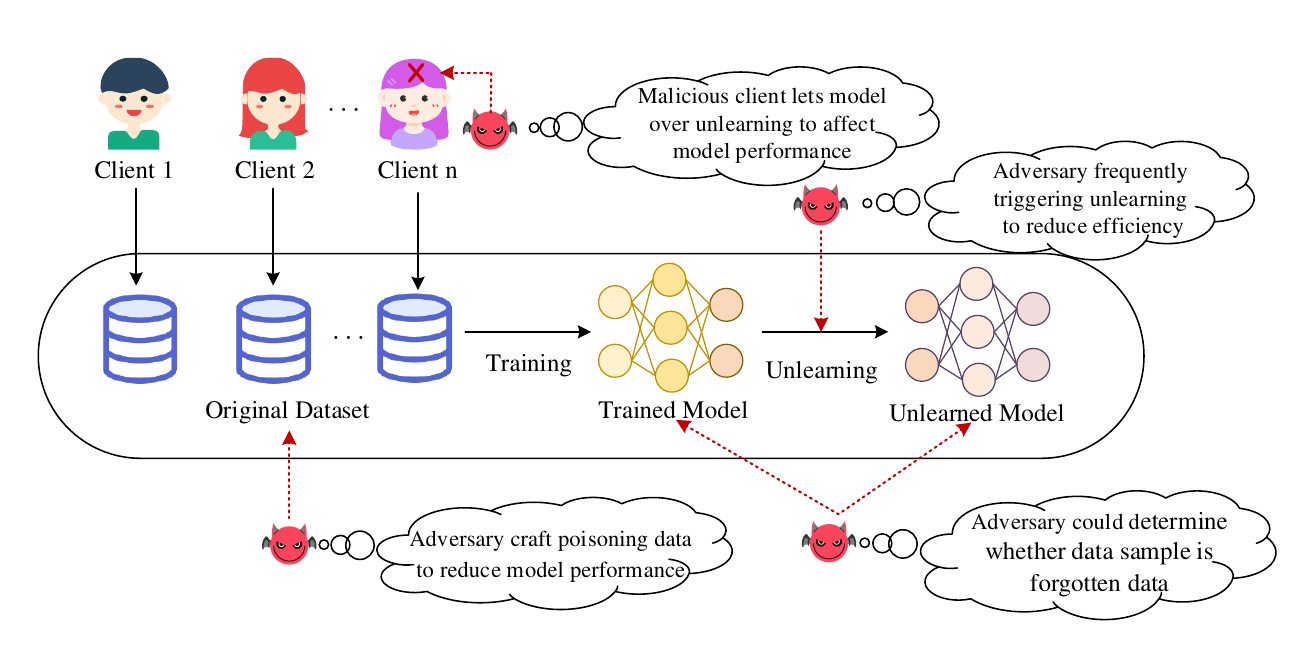}
    \caption{Purpose of Attacks against Machine Unlearning.}
    \label{fig:attack}
\end{figure*}

\end{itemize}
\subsubsection{Active Defense}
Before the model suffers from various privacy attacks, machine unlearning can proactively eliminate sensitive information related to the user's private data in the model. As a result, adversaries are unable to ascertain whether the forgotten data was part of the training dataset, obtain the associated attribute information, or attempt to reconstruct the data.

\begin{itemize}
    \item {\textbf{Defense of Membership Inference Attack.}
     Membership inference attacks aim at determining whether specific data points are present in the training dataset. The adversary exploits the differences in the target model's behavior (output probabilities or confidence scores) on training and non-training data to perform membership inference.
    
    Studies \cite{Graves21,Golatkar20Forgetting,Golatkar21,Chundawat23Zero-Shot} eliminate the impact of the target data in the model by using different machine unlearning algorithms (detailed in Section \ref{3}), thus effectively preventing this attack. After successful unlearning, the model holds no sensitive information about the target data, preventing attacks.}
    \item {\textbf{Defense of Property Inference Attack.} 
    Ganju \textit{et al.} \cite{Ganju18} introduced the property inference attack, which is intended to extract statistical properties of the underlying training data in machine learning models. Specifically, the adversary typically exploits patterns and correlations within accessible data to infer unknown sensitive properties.

     Stock \textit{et al.} \cite{Stock23} proposed first property unlearning as an effective defense against white-box property inference attacks. This method systematically alters the trained weights and biases in a target model to prevent an adversary from inferring properties. }
     
    \item {\textbf{Defense of Model Inversion Attack.}
    The adversary of model inversion attacks uses access to an ML model to reconstruct sensitive details about the training data. For instance, they might carefully analyze the model's outputs to reconstruct features of the original training data.

    As described in \cite{Graves21,Chundawat23Zero-Shot} (detailed in Section \ref{3}), machine unlearning eliminates private information about the target data in the model. Therefore, the model that completes the unlearning eliminates traces of the target data.}
\end{itemize}

\vspace{0.15cm}
\noindent\textbf{Summary:} In addition to enhancing data privacy, machine unlearning has significant potential in a variety of applications. Beyond the applications mentioned above, it can be used in pre-trained generative adversarial networks to prevent the generation of outputs containing undesirable features \cite{Tiwary2023Generative}. 
Additionally, it is useful in machine learning-based access control system management \cite{Javier2023Machine} and can be applied to concept drift \cite{ArteltMPPH23Drift} without the need for retraining. 
It proves invaluable in medical classification to mitigate bias \cite{BevanA22medical}, as well as in lifetime anomaly detection \cite{Du19Lifelong} and causal inference \cite{Ramachandra2023Causal}. Moreover, it serves the purpose of identifying critical and valuable data samples within a model, while helping to address fairness issues.

\begin{table*}[!t]
\centering
\caption{Attacks against Machine Unlearning}
\label{table:attack}
\scalebox{0.65}{
\renewcommand{\arraystretch}{1.5}
\begin{tabular}{|l|c|c|l|l|}
\hline
 \textbf{Attack} &\textbf{Paper}   &\textbf{Assumption}  & \textbf{Aim}    &\textbf{Limitation} \\ \hline
\multirow{2}{*}[-2.5ex]{Membership inference attack}   & \cite{Chen21When}          &   Black-box    
& \begin{tabular}[c]{@{}l@{}}Determine whether a data sample is forgotten\end{tabular} 
& \begin{tabular}[c]{@{}l@{}}Can be defensed by temperature scaling, differential privacy, \\publishing the label only or the top-k confidence values\end{tabular}         \\
\cline{2-5}
&\cite{Lu22}  
& Black-box              
& \begin{tabular}[c]{@{}l@{}}Determine whether a data sample is forgotten\end{tabular} 
& \begin{tabular}[c]{@{}l@{}}Computationally expensive,\\Not effective for the latest unlearning methods \cite{Mehta22,Kim22}\end{tabular}          \\\hline

\multirow{2}{*}{Data poising attack}   & \cite{Marchant22}  
&  \begin{tabular}[c]{@{}l@{}}White-box \\ Grey-box\end{tabular}       
&\begin{tabular}[c]{@{}l@{}}Reduce the efficiency of unlearning process \end{tabular}      & \begin{tabular}[c]{@{}l@{}} This attack only target at certified unlearning\cite{Guo20,Neel21} \end{tabular}        \\
\cline{2-5}
& \cite{Di22}   
&  Grey-box   
&\begin{tabular}[c]{@{}l@{}}Reduced prediction accuracy for specific samples\end{tabular} 
& \begin{tabular}[c]{@{}l@{}}This attack assumes many premises that are impractical\end{tabular} \\  
\hline
Over-unlearning attack &\cite{Hu23}
&Black-box
&\begin{tabular}[c]{@{}l@{}}Reduce the predictive  performance of the model\end{tabular} 
& \begin{tabular}[c]{@{}l@{}}Mainly target at gradient-based  approximate unlearning\end{tabular}  \\  
\hline                                               
\end{tabular}
}
\end{table*}

\section{Attacks on Machine Unlearning}
\label{7}
While the dynamic nature of machine unlearning safeguards the data contributor's privacy, it might inadvertently expose traces of forgotten data. This could potentially offer adversaries new avenues for attack, thus making the unlearned model more vulnerable than the original model \cite{Di22}. For instance, excessive unlearning could lead to significant parameter shifts, a phenomenon known as the `Streisand effect'. Adversaries may keenly detect these shifts, triggering potential privacy breaches. This runs counter to machine unlearning's basic design philosophy of protecting privacy. As mentioned in Section \ref{2}, unlearning algorithms have four properties, different attacks on unlearned models jeopardize these properties. For example, membership inference attacks threaten private data and compromise the unlearning effectiveness. Data poisoning attacks can lead to high latency and computational overhead, thereby undermining the unlearning efficiency. Furthermore, both data poisoning and over-unlearning attacks have the potential to diminish unlearned model utility.

\subsection{MU-specific Membership Inference Attack}
\subsubsection{Threat Model}

\begin{itemize}
\item{\textbf{Adversary's Goal.} Because $\mathcal{D}_u$ often tends to contain more valuable sensitive information than $\mathcal{D}_r$, the goal of a membership inference attack is to determine whether a target sample is forgotten data \cite{Chen21When}. More generally, as shown in Fig.~\ref{fig:attack}, the adversary aims to infer whether the target sample $x_{i}$ belongs to $\mathcal{D}_u$ but not in $\mathcal{D}_r$ \cite{Lu22,Kurmanji23}.}

\item{\textbf{Assumptions.} Assuming the adversary lacks knowledge of the model's internal structure but has black-box access to both the target original and unlearned models \cite{Chen21When}. Additionally, the adversary possesses a local shadow dataset that can be utilized to train a multitude of shadow models imitating the target model's behavior. Then, shadow models can be employed to produce training meta-data for the attack meta-model. \cite{Lu22,Kurmanji23}.}
\end{itemize}

\subsubsection{Attack Methods} In 2021, Chen \textit{et al.} \cite{Chen21When} first investigated unintentional privacy leakage caused by machine unlearning, proposing a new membership inference attack that leverages the different outputs (posteriors) of the models before and after unlearning to determine if the target sample is in $\mathcal{D}_u$. They also pointed out that temperature scaling, only releasing predictive labels, and differential privacy could effectively defend against this attack. Subsequently, in 2022, Lu \textit{et al.} \cite{Lu22} further highlighted that even unlearned models releasing only predictive labels are still vulnerable, and proposed a membership inference attack independent of posterior probabilities. They observed differential predictions between the original and unlearned models by injecting perturbations into the target samples. These differential predictions could then be used to infer whether samples were part of the $\mathcal{D}_u$. However, the computational cost of this attack is expensive.

\subsection{MU-specific Data Poisoning Attack}
\subsubsection{Threat Model}
\begin{itemize}
\item {\textbf{Adversary's Goal.} Two purposes could be achieved through the data poisoning attack. Firstly, the adversary aims to decrease the promised unlearning efficiency gains from model providers by frequently triggering the unlearning process \cite{Marchant22}. Secondly, the purpose is to make the unlearned model incorrectly classify data, reducing prediction accuracy for specific samples \cite{Di22}.}
\item {\textbf{Assumptions.} For different purposes, attacks are based on different assumptions. For the first purpose, assume the adversary possesses both white-box and grey-box access to the target model \cite{Marchant22}. Under the white-box setting, the adversary has the ability to access the model structure and the state, the training data of benign users. Under the grey-box setting, the adversary only knows the model architecture \cite{Marchant22}. For the second purpose, assume that the adversary has grey-box access, which could access to target model architecture and gradients \cite{Di22}. }
\end{itemize}

\subsubsection{Attack Methods} Marchant \textit{et al.} \cite{Marchant22} proposed the first data poisoning attack targeting unlearning efficiency, known as the slow-down attack. This attack crafts poisoning schemes via careful noise addition to triggering the unlearning process far more than typically required, significantly increasing the computational and time consumption, similar to traditional denial-of-service attacks. Di \textit{et al.} \cite{Di22} attempted to reduce the predictive performance of the model for specific samples. They created camouflage data points to mask the negative impacts of the poisoned dataset, making the unlearned model misclassify the target test point, thus achieving targeted poisoning attacks. 

\subsection{Over-unlearning Attack}
\subsubsection{Threat Model}
\begin{itemize}
\item {\textbf{Adversary's Goal.} In MLaaS, the adversary (malicious user) has the potential to compromise the unlearned model's utility. They can request a model provider to unlearn specially crafted data to make the model unlearn more information than expected (normal unlearn requests), thus achieving the goal of reducing the predictive performance of the model \cite{Hu23}. }

\item {\textbf{Assumptions.} The adversary only has black-box access to the model and the unlearning process takes place on the server side \cite{Hu23}.}
\end{itemize}

\subsubsection{Attack Methods} Hu \textit{et al.} \cite{Hu23} proposed an over-unlearning attack, which involves blending extra samples (as the crafted data) from a different task into the $\mathcal{D}_u$. As the model provider attempts to eliminate the crafted data, the model inadvertently discards extra information related to the other task. This results in over unlearning, thereby diminishing the prediction performance of the unlearned model. They also pointed out a fundamental difference between the over-unlearning attack and the data poisoning attack. Although similar, over-unlearning targets approximate unlearning under a blending scenario that does not require the inclusion of a training or re-training procedure, which is typically critical for the data poisoning attack.

\vspace{0.15cm}
\noindent\textbf{Summary:} While attacks against machine unlearning are relatively less prevalent so far, it is important to consider the security of the unlearning process. 
These attacks can serve as verification metrics to assess the effectiveness of machine unlearning algorithms and guide the construction of better algorithms to circumvent these vulnerabilities. Table \ref{table:attack} summarizes existing attacks against machine unlearning possess limitations. Some attacks require stringent conditions, some have a narrow scope of applicability, and some can be mitigated by recently proposed machine unlearning algorithms.

\begin{figure*}[htbp]
    \centering
    \includegraphics[width=0.9\linewidth]{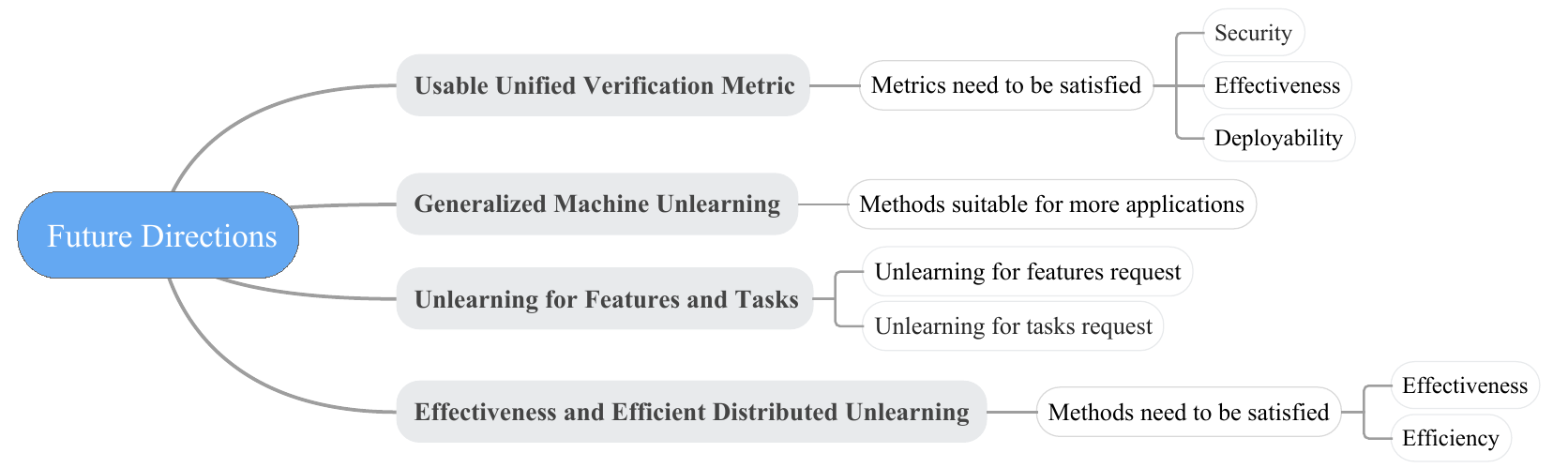}
    \caption{Future Directions.}
    \label{fig:directions}
    \end{figure*}

\section{Challenges and Prospects}
\label{8}

Here, we highlight the challenges of machine unlearning and prospect promising future directions that could serve as a beacon for innovative research.  
\subsection{Usable Unified Verification Metric}
Currently, there is no unified standard for verifying the quality of unlearning, and we perceive that the existing verification metrics are inadequate for the following reasons:
\begin{itemize}
    \item Verification metrics should not have any negative impact on the model, whereas some metrics impair the predictive capacity of the model and pose potential privacy threats. For instance, watermark-based metrics often embed backdoors in the model, undermining its predictive accuracy \cite{Sommer20, gao2022verifi}.
    \item Then, verification metrics should be easy to understand and deploy for the ordinary user, and many metrics fall short of this requirement. For example, cryptographic-based metric \cite{Eisenhofer2022Verifiable} presents a challenge to non-security users as the verification process is not easy to understand for them.
\end{itemize}

Therefore, designing a secure, effective, easy-to-implement, and understandable verification metric as a unified standard is worth careful consideration.

\subsection{Generalized Machine Unlearning}

Traditional machine unlearning is not applied to practical applications \cite{Kurmanji23}. This is mainly because it fails to consider the unique connections among data, which could disrupt the model structure. Moreover, according to various applications, unlearning should have different objectives and priorities \cite{Kurmanji23}. For instance, in privacy-focused applications, the primary objective is privacy protection, and it is acceptable to sacrifice the model's performance to some extent. In contrast, when the model needs to remove outdated data, the main objective is to maintain the model's performance, thus leaving behind some related information that is insignificant \cite{Kurmanji23}.  Furthermore, the majority of the existing unlearning methods target classification tasks. There is a dearth of studies concerning regression and generative tasks, impeding the broader application of machine unlearning. Therefore, seeking generalized machine unlearning that applies to different real-world applications is critically important.

\subsection{Effectiveness and Efficiency Distributed Unlearning}
Presently, the study on distributed unlearning is limited. First, the distributed unlearning study mainly focuses on FL settings. However, other distributed machine learning settings, such as split learning, collaborative learning, and peer-to-peer learning, also need to unlearn data to satisfy `the right to be forgotten' and user demands, to enhance model robustness and privacy. Second, current federated unlearning methods fail to simultaneously satisfy effectiveness and efficiency. For example, some methods implement unlearning on the server side, and sensitive information about the unlearned data still remains in the global model \cite{Liu21FedEraser,Wu22Federated}. Additionally, some methods necessitate interaction between the server and clients during the unlearning process, resulting in excessive consumption of time \cite{Liu22right,Wang23BFU,Che23,Zhu23,Zhang22Prompt}. Therefore, it is worth exploring the implementation of effectiveness and efficiency unlearning in various distributed learning settings.

\subsection{Unlearning for Features and Tasks}
Unlearning algorithms on date focus on class-based and sample-based requests, which fall short of meeting user demands. First, privacy leaks can originate from datasets that share particular features \cite{Warnecke23}. Consider a credit assessment service, where it's essential to unlearn specific features like marital status or religious beliefs of applicants to prevent bias. Moreover, machine learning models are trained not only for one task but for multiple tasks \cite{Liu2022}. In such settings, it is necessary to remove private data related to a specific task \cite{Liu2022}. For example, imagine an AI tutor designed for personalized academic assistance. After a student's course completion, the AI tutor may need to unlearn the personalized teaching strategies tailored to that student. 
In both situations, sequentially unlearning the samples is inadvisable due to the high computational cost and potential decrease in model utility. Consequently, unlearning at the feature or task level is essential to address diverse real-world requirements.

\vspace{0.15cm}
\noindent\textbf{Summary:} Machine unlearning is in its early stages of development and faces various challenges that need to be addressed. We believe that the directions outlined above (as shown in Fig.~\ref{fig:directions}) hold promising potential for future research and will bring unprecedented advancements. We hope that these insights will inspire scholars in their ongoing exploration. 

\section{Conclusion}
In recent years, numerous legal regulations have emerged requiring model (service) providers to promptly and effectively delete users' data and its impact on models in response to their requests. 
Machine unlearning is a new technology that can satisfy such as `the right to be forgotten'. 
In this survey, we have presented a comprehensive introduction to machine unlearning, providing the basic knowledge for interested scholars. 
To ensure the practical implementation of MU, we have differentiated between verification and evaluation metrics, systematically summarizing and categorizing each respectively.
Moreover, in categories of exact unlearning and approximate unlearning, we provide a more sub-categorization based on different underlying strategies used. 
Additionally, our survey highlights the analysis of MU within the distributed learning setting, particularly, the focused federated learning.
Furthermore, we have pointed out the significant potential of MU applications and summarized specific attacks against MU. Finally, we have underscored existing challenges and prospected potential future directions worth exploring. 

\bibliographystyle{ACM-Reference-Format}
\bibliography{sample-base}

\end{document}